\documentclass[10pt,twocolumn,letterpaper]{article}

\usepackage{cvpr}              %
\usepackage[accsupp]{axessibility} 

\usepackage[utf8]{inputenc} %
\usepackage[T1]{fontenc}    %
\usepackage{url}            %
\usepackage{booktabs}       %
\usepackage{amsfonts}       %
\usepackage{nicefrac}       %
\usepackage{microtype}      %
\usepackage[dvipsnames]{xcolor}
\usepackage{graphicx}
\usepackage[ruled,vlined]{algorithm2e}
\usepackage{wrapfig}
\usepackage{subcaption}
\usepackage{xspace}
\usepackage{multirow}
\usepackage{amssymb}%
\usepackage{amsmath}
\usepackage{pifont}%
\usepackage{graphbox} %
\usepackage{enumitem}

\usepackage[pagebackref=true,breaklinks=true,colorlinks,bookmarks=false]{hyperref}

\newcommand{\calP}{\mathcal{P}}
\newcommand{\calB}{\mathcal{B}}
\newcommand{\calD}{\mathcal{D}}
\newcommand{\calQ}{\mathcal{Q}}
\newcommand{\calN}{\mathcal{N}}
\newcommand{\calS}{\mathcal{S}}

\def\eg{\textit{e.g.}\@\xspace}
\def\ie{\textit{i.e.}\@\xspace}

\definecolor{mygray}{gray}{0.4}
\newcommand{\cmark}{\color{mygray}\ding{51}}%
\newcommand{\xmark}{\color{mygray}\ding{55}}%

\title{Bongard-HOI: Benchmarking Few-Shot Visual Reasoning \\for Human-Object Interactions}

\author{
Huaizu Jiang$^{1*}$, Xiaojian Ma$^2$\thanks{First two authors contributed equally.} , Weili Nie$^3$ , Zhiding Yu$^3$ , \\Yuke Zhu$^{3,4}$, Song-Chun Zhu$^2$, Anima Anandkumar$^{3,5}$  \\
$^1$Northeastern University~~~$^2$UCLA~~~$^3$NVIDIA~~~$^4$UT Austin~~~$^5$Caltech \\
\texttt{\small{h.jiang@northeastern.edu, xiaojian.ma@ucla.edu, \{wnie,zhidingy\}@nvidia.com},}\\
\texttt{\small{yukez@cs.utexas.edu, sczhu@stat.ucla.edu, anima@caltech.edu}}}

\def\cvprPaperID{1795} %
\def\confName{CVPR}
\def\confYear{2022}

\begin{document}

\maketitle

\begin{abstract}
A significant gap remains between today's visual pattern recognition models and human-level visual cognition especially when it comes to few-shot learning and compositional reasoning of novel concepts. We introduce \textbf{Bongard-HOI}, a new visual reasoning benchmark that focuses on compositional learning of human-object interactions (HOIs) from natural images. It is inspired by two desirable characteristics from the classical Bongard problems (BPs):  1) few-shot concept learning, and 2) context-dependent reasoning. We carefully curate the few-shot instances with hard negatives, where positive and negative images only disagree on action labels, making mere recognition of object categories insufficient to complete our benchmarks. We also design multiple test sets to systematically study the generalization of visual learning models, where we vary the overlap of the HOI concepts between the training and test sets of few-shot instances, from partial to no overlaps. Bongard-HOI presents a substantial challenge to today's visual recognition models. The state-of-the-art HOI detection model achieves only 62\% accuracy on few-shot binary prediction while even amateur human testers on MTurk have 91\% accuracy. With the Bongard-HOI benchmark, we hope to further advance research efforts in visual reasoning, especially in holistic perception-reasoning systems and better representation learning. Code
is available.\footnote{\href{https://github.com/nvlabs/Bongard-HOI}{https://github.com/nvlabs/Bongard-HOI}}
\end{abstract} 

\begin{figure}[t]
\centering
\renewcommand{\tabcolsep}{1.5pt}
\newcommand{\loadFig}[1]{\includegraphics[height=0.245\linewidth]{#1}}
\begin{tabular}{cc|cc}
\multicolumn{2}{c|}{positive examples} & \multicolumn{2}{c}{negative examples} \\
\multicolumn{2}{c|}{\textbf{\texttt{ride bicycle}}} & 
\multicolumn{2}{c}{\textbf{\texttt{!ride bicycle}}} \\
\loadFig{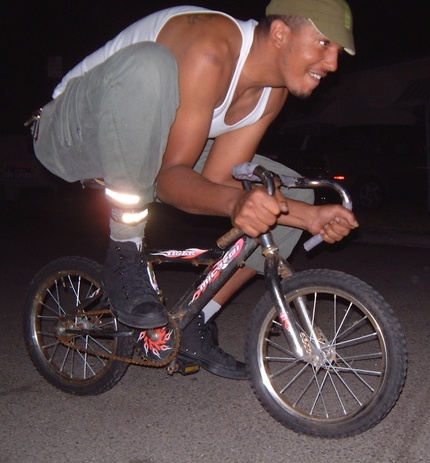} & 
\loadFig{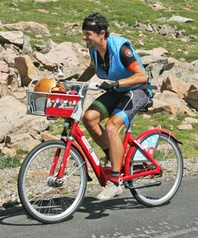} & 
\loadFig{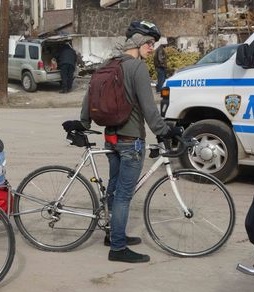} & 
\loadFig{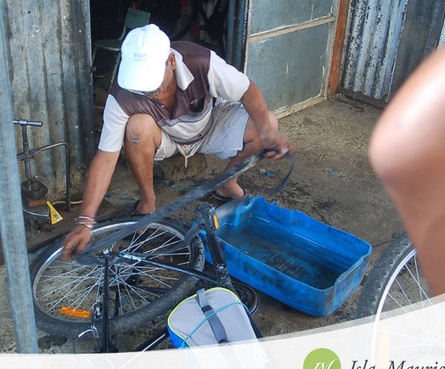} \\
\loadFig{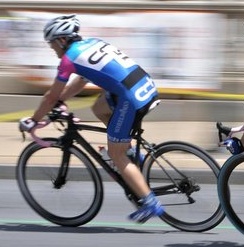} & 
\loadFig{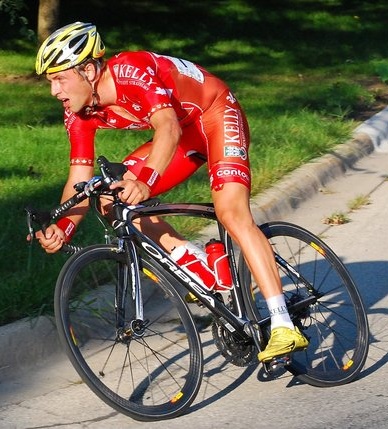} & 
\loadFig{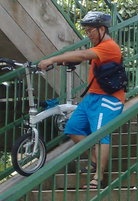} & 
\loadFig{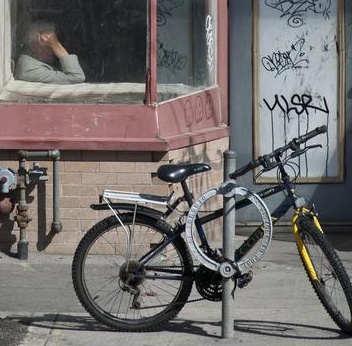} \\ 
\midrule
\multicolumn{4}{c}{Query images:} \\
\multicolumn{2}{r}{\loadFig{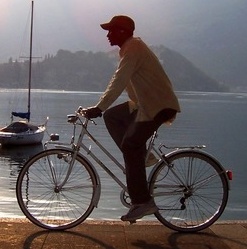}} &
\multicolumn{2}{l}{\loadFig{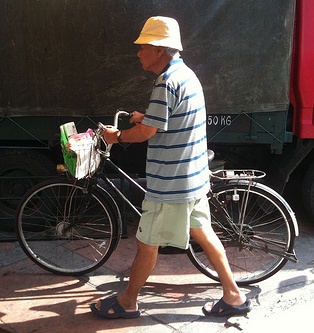}} \\
Labels: & \multicolumn{1}{l}{\textbf{positive}} & \multicolumn{2}{l}{\textbf{negative}} \\
\end{tabular}
\vskip -0.1in
\caption{\textbf{Illustration of a few-shot learning instance from our Bongard-HOI benchmark.} The positive images in the top left part follow the visual relationship of riding a bike between the person and objects while such a relationship does not exist in the negative examples.
Note that an actual problem in Bongard-HOI contains 6 images of positive examples, 6 negative examples, and 1 query image, which is different from the illustration here.
}
\label{fig:teaser}
\vskip -0.2in
\end{figure}

\section{Introduction}
In recent years, great strides have been made on visual recognition benchmarks, such as ImageNet~\cite{deng2009imagenet} and COCO~\cite{patterson2016coco}. Nonetheless, there remains a considerable gap between machine-level pattern recognition and human-level cognitive reasoning. Current image understanding models typically require a large amount of training data yet struggle to generalize beyond the visual concepts seen during training. In contrast, humans can reason about new visual concepts in a compositional manner from just a few examples~\cite{lake2015human}. To march towards human-level visual cognition, we need to depart from conventional benchmarks on closed-vocabulary recognition tasks and aim to systematically examine compositional and few-shot learning of novel visual concepts.

While existing benchmarks such as miniImageNet~\cite{vinyals2016matching}, Meta-Dataset~\cite{triantafillou2019meta}, and ORBIT~\cite{triantafillou2019meta} have been dedicated to studying few-shot visual learning, they focus on recognizing object categories instead of the compositional structures of visual concepts, \eg, visual relationships. A parallel line of research aims at building benchmarks for abstract reasoning by taking inspiration from cognitive science such as  RPM (Raven-style Progressive Matrices)~\cite{barrett2018measuring,teney2019v} and Bongard-LOGO~\cite{bongard1968recognition,nie2020bongard}. In these benchmarks, a model has to learn concept induction rules from a few examples and the concepts are context-dependent in each task. However, they use 
simple synthetic images~\cite{barrett2018measuring,nie2020bongard} or focus on basic object-level properties, such as shapes and categories~\cite{teney2019v}.

\noindent\textbf{Our new benchmark:} In this paper, we introduce \textbf{Bongard-HOI}, a new benchmark for compositional visual reasoning with natural images. It studies human-object interactions (HOIs) as the visual concepts, requiring explicit compositional reasoning of object-level concepts. Our Bongard-HOI benchmark inherits two important characteristics of the classic Bongard problems (BPs)~\cite{bongard1968recognition}: 1) \emph{few-shot binary prediction}, where a visual concept needs to be induced from just six positive and six negative examples and 2) \emph{context-dependent reasoning}, where the label of an image may be interpreted differently under different contexts.

\begin{figure}[t!]
\renewcommand{\tabcolsep}{1.5pt}
\newcommand{\loadFig}[1]{\includegraphics[height=0.275\linewidth]{#1}}
\begin{tabular}{ccc}
\multicolumn{3}{c}{\textbf{actions with dogs}} \\
\loadFig{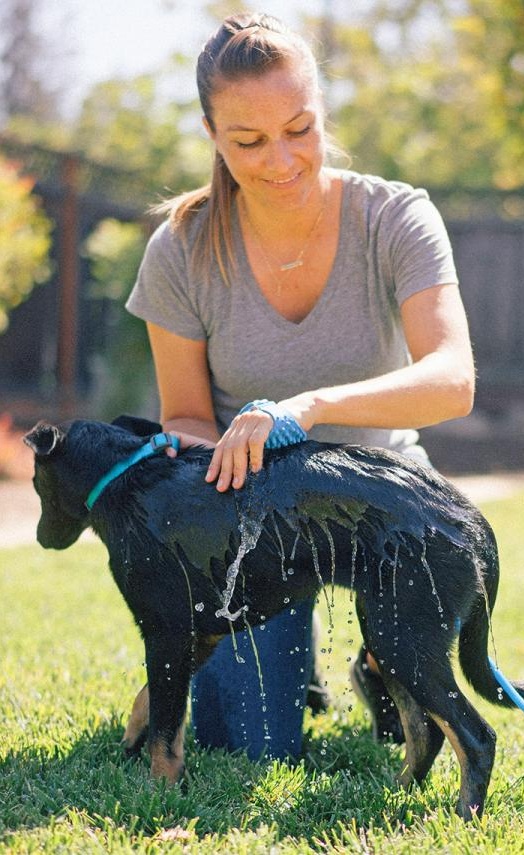} & 
\loadFig{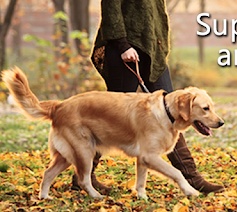} & 
\loadFig{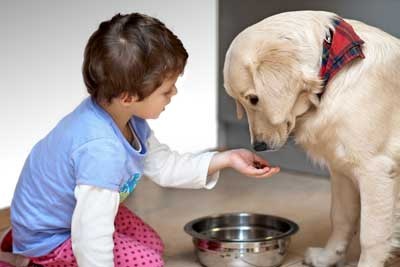} \\
\multicolumn{3}{c}{\textbf{actions with oranges}} \\
\loadFig{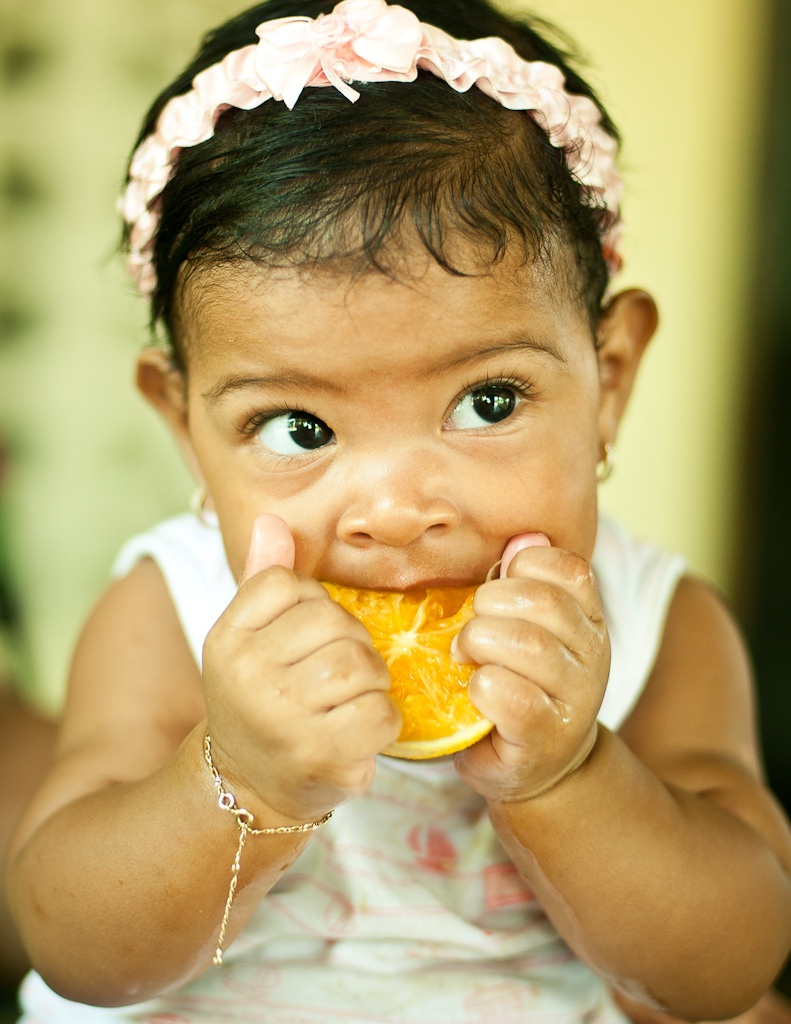} & 
\loadFig{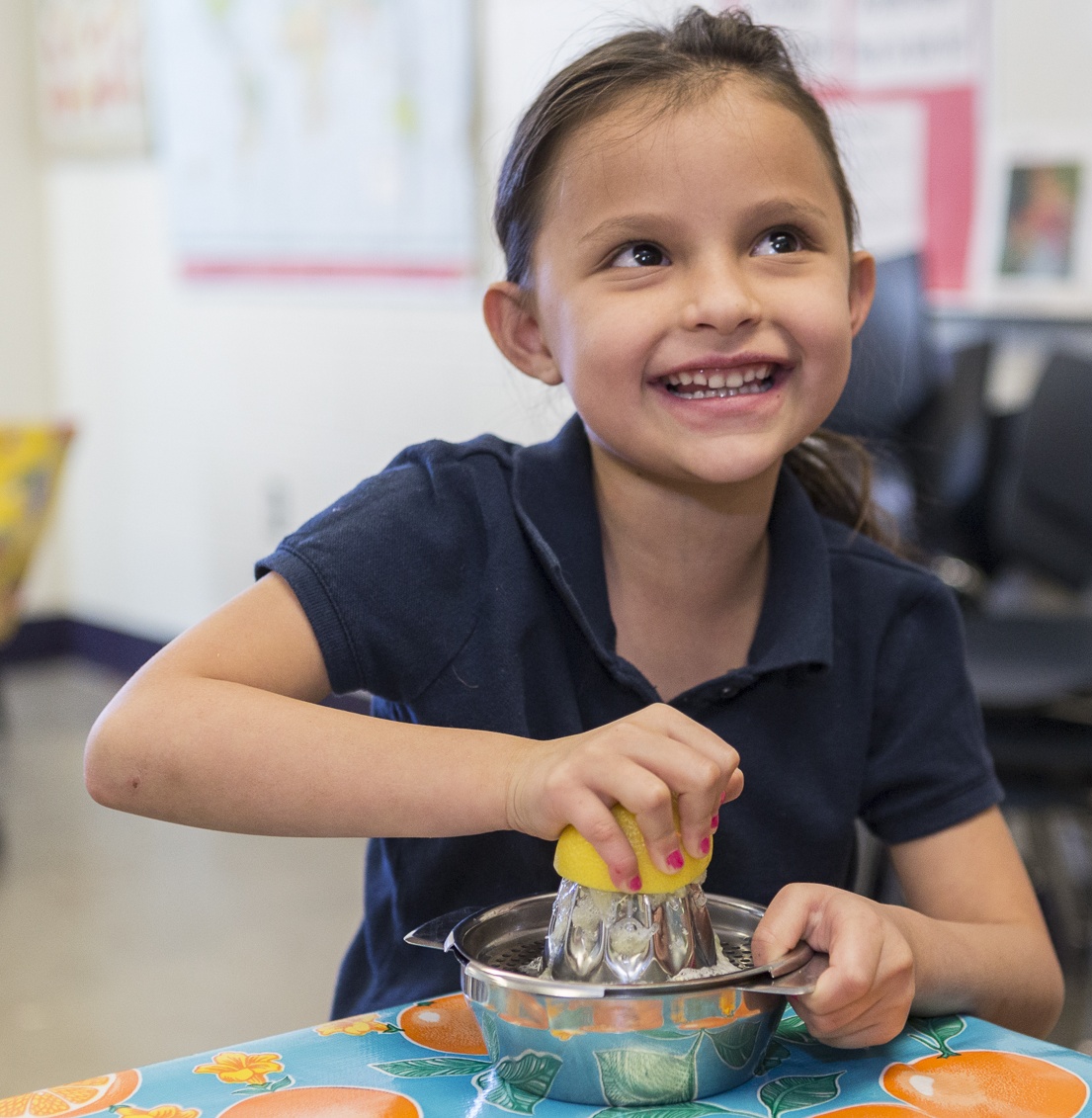} & 
\loadFig{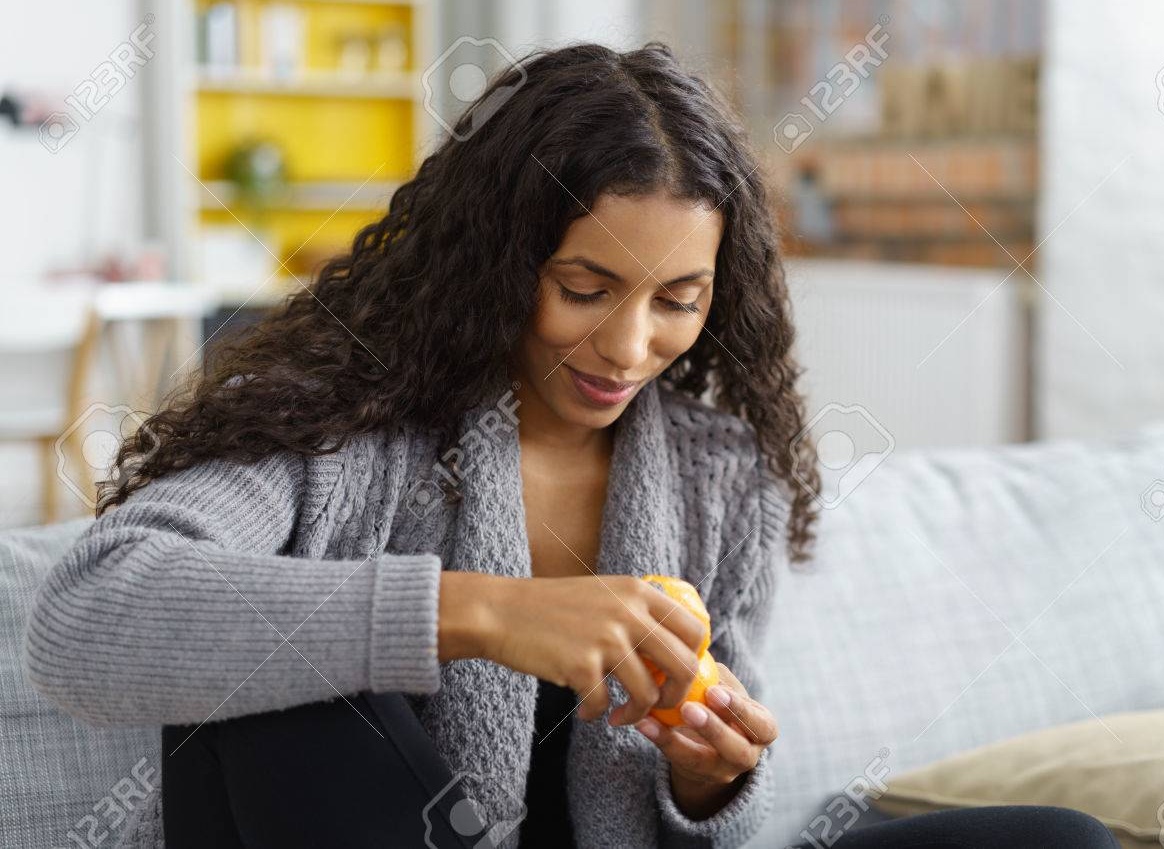} \\
\end{tabular}
\vskip -0.1in
\caption{\textbf{Examples of different actions with the same object.} From top to bottom, left to right: washing, walking, and feeding dogs; eating, squeezing, and peeling oranges. To differentiate these images, we need compositional understanding on both the actions and the objects. We exploit this to select \textit{hard negatives} in Bongard-HOI: negative images contain the same object as the positives, but the actions are different.}
\label{fig:hard_negatives}
\vskip -0.2in
\end{figure}

Furthermore, Bongard-HOI upgrades the original BPs from synthetic graphics to natural images. Our benchmark contains rich visual stimuli featuring large intra-class variance, cluttered background, diverse scene layouts, etc. In Bongard-HOI, a single few-shot binary prediction instance, referred to as BP, contains a set of six positive images and a set of six negative images, along with query images (see Fig.~\ref{fig:teaser} for examples). The task is making binary predictions on the query images.

We construct the few-shot instances in Bongard-HOI on top of the HAKE dataset~\cite{li2019hake,li2020detailed}.
To encourage the explicit reasoning of visual relationships, we use \emph{hard negatives} to construct few-shot instances. The hard negatives consist of negatives that contain
objects from the same categories as those contained in the positive sets but with different action labels. Fig.~\ref{fig:hard_negatives} presents some examples of these images. Since both positive and negative examples contain object instances from the same categories, mere recognition of object categories is insufficient to complete the tasks. 
Rather, reasoning about visual relationships between person and objects is required to solve these few-shot binary prediction problems. The existence of such hard negatives distinguishes our benchmark from existing visual abstract reasoning counterparts~\cite{barrett2018measuring,teney2019v,nie2020bongard}. Comparisons with different benchmarks can be found in Table~\ref{tab:dataset_comparison}. 

\begin{figure*}[t!]
\centering
\renewcommand{\tabcolsep}{1.5pt}
\newcommand{\loadFig}[1]{\includegraphics[align=c, height=0.15\linewidth]{#1}}
\newcommand{\seen}[1]{\textcolor{OrangeRed}{#1}}
\newcommand{\unseen}[1]{\textcolor{RoyalBlue}{#1}}
\begin{tabular}{cccccc}
     & 
    \seen{\textbf{\texttt{sit\_on bed}}} & 
    \seen{\textbf{\texttt{straddle bicycle}}} &
    \seen{\textbf{\texttt{hug person}}} &
    \seen{\textbf{\texttt{wash car}}} &
     \\
    \textbf{training set} & 
    \loadFig{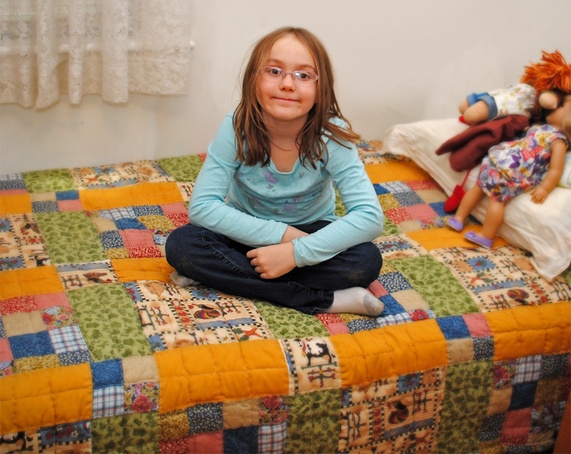} &
    \loadFig{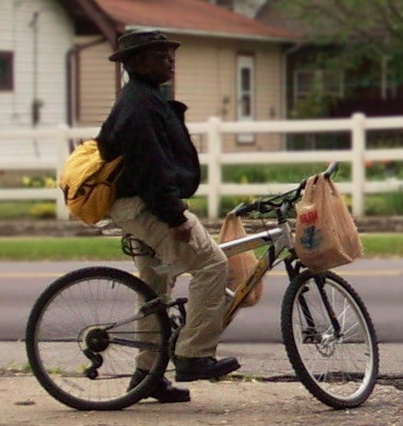} &
    \loadFig{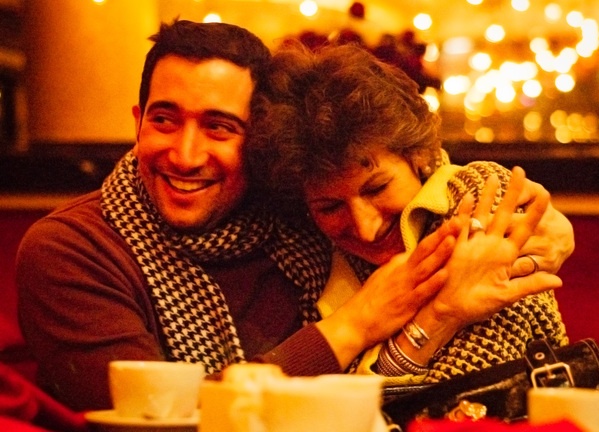} & 
    \loadFig{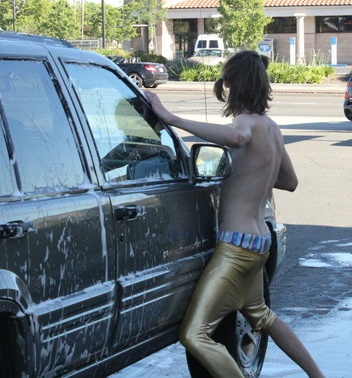} & 
    $\mathbf{\cdots}$ \\
    \midrule
     &
     \seen{\textbf{\texttt{wash bicyle}}} &
     \seen{\textbf{\texttt{sit\_on}}}~~~\unseen{\textbf{\texttt{bench}}} &
     \unseen{\textbf{\texttt{greet}}}~~~\seen{\textbf{\texttt{person}}} &
     \unseen{\textbf{\texttt{shear sheep}}} &
     \\
    \textbf{test set} &
    \loadFig{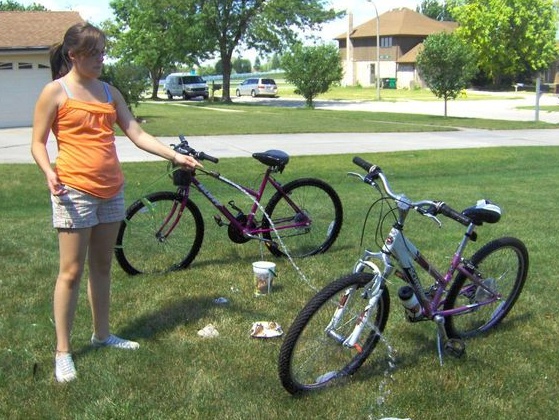} &
    \loadFig{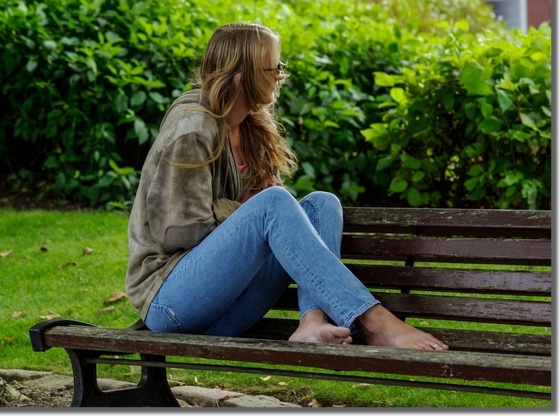} &
    \loadFig{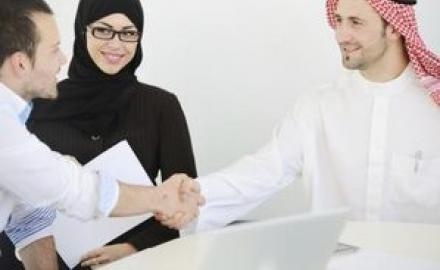} & 
    \loadFig{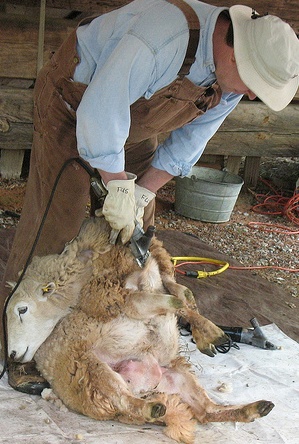} & 
     \\
     &
    \textbf{seen act. seen obj.} &
    \textbf{ seen act. unseen obj.} &
    \textbf{unseen act. seen obj.} &
    \textbf{ unseen act. unseen obj.} &\\
\end{tabular}
\vskip -0.1in
\caption{\textbf{Illustration of our four separate test sets for different types of generalization.} We show a few HOI concepts in the training and test sets in the top and bottom row, respectively. We use the \seen{\textbf{\texttt{red}}} fonts to denote an object or action class that is available in the training set and \unseen{\textbf{\texttt{blue}}} fonts indicate those held-on unseen ones in the test set.}
\label{fig:generalization}
\vskip -0.2in
\end{figure*}

We carefully curate the annotations in HAKE when constructing the few-shot instances. Recall the visual concept contained in the positive images should not appear in any of the negative ones. Thus, we have to carefully select the images in both sets.
We employ high-quality annotators from the Amazon Mechanical Turk platform to curate the test set to further remove ambiguously and wrongly labeled few-shot instances. 
In this process, 2.5\% of the few-shot instances in the test set are discarded. We end up with 23K and 15K few-shot instances in disjoint training and test sets, respectively. %

An important goal of the Bongard-HOI benchmark is to \textit{systematically} study the generalization of machine learning models for real-world visual relationship reasoning. To this end, we introduce four separate test sets to investigate different types of generalization, depending on whether the action and object classes are seen in the training set. Fig.~\ref{fig:generalization} illustrates their design.
This way, we have full control of the overlap between the concepts (\ie, HOIs) between training and test of few-shot instances. It enables us to carefully examine the generalization of visual learning models. Ideally, a learning model should be able to generalize beyond the concepts it has seen during training. 
Even for unseen HOI concepts, 
the model should be able to learn \emph{how to induce} the underlying visual relationship from just a few examples. 

\noindent\textbf{Establishing baselines}: In our experiments, we first examine the state-of-the-art HOI detection models' performance on this new task, we trained an oracle model with HOITrans~\cite{zou2021_hoitrans} on all the HOI categories, \emph{including those in the test sets of our Bongard-HOI benchmark}, and output binary prediction on the query image via a majority vote based on HOI detections. Its accuracy is only 62.46\% (with a chance performance of 50\%), demonstrating the challenge of our visual reasoning tasks. We then evaluate state-of-the-art few-shot learning approaches, including non-episodic and meta-learning methods. We show that the current learning models struggle to solve the Bongard-HOI problems. Compared to amateur human testers' 91.42\% overall accuracy, who have access to a few examples of visual relationships before working on solving our problems, the state-of-the-art few-shot learning model~\cite{chen2020new} only has 55.82\% accuracy.

The results above lead to this question: \textit{why do they perform so poorly?} To this end, we offer a detailed analysis of the results and propose several conjectures. The first one is a lack of holistic perception and reasoning systems, since models that have only good pattern recognition performances, \eg HOITrans, are likely to fail on our benchmarks. Moreover, we believe there is a need for additional representation learning, \eg pre-training, since currently we only train on binary labels of few-shot instances. Nonetheless, we believe much effort is still needed to further investigate the challenges brought by our benchmark.

To sum up, this paper makes the following contributions:
\setlength{\leftmargini}{0.85em}
\begin{itemize}[topsep=0pt]
    \item We introduce Bongard-HOI, a new benchmark for few-shot visual reasoning with human-object interactions, aiming at combining the best of few-shot learning, compositional reasoning, and challenging real-world scenes.
    \item We carefully curate Bonagrd-HOI with hard negatives, making mere recognition of object categories insufficient to complete our tasks. We also introduce multiple test sets to systematically study different types of generalization.
    \item We analyze state-of-the-art few-shot learning and HOI detection methods. However, experimental results show their inability on achieving good results on Bongard-HOI. Our conjectures suggest future research in models with holistic perception-reasoning systems and better representations.
\end{itemize}

\begin{table*}[t!]
\centering
\caption{\textbf{An overview of different benchmark datasets covering HOI detection, few-shot learning, and abstract visual reasoning.} In the first row, the abbreviation \textit{ctx} denotes context; \textit{generalization types} indicates if a benchmark includes multiple test splits to examine different types of generalization. $^*$We consider the concept of object counts as compositional while others such as object attributes and categories not~\cite{teney2019v}).} 
\label{tab:dataset_comparison}
\vskip -0.1in
\renewcommand{\tabcolsep}{2pt}
\small
\begin{tabular}{c|c|c|c|c|c|c|c|c|c}
\toprule
 & \multirow{2}{*}{concept} & compositional & ~natural~ & ~few-~ & ctx-dependent & hard & generalization & \multirow{2}{*}{\#concepts} & \multirow{2}{*}{\#tasks} \\
 & & concept & image & shot & reasoning & negatives & types &  & \\
\midrule
HAKE~\cite{li2019hake,li2020detailed} & HOI & \cmark & \cmark & \xmark & \xmark & \cmark & \xmark & 600 & 122.6K \\ %
\midrule
Omniglot~\cite{lake2011one} & shape & \xmark & \xmark & \cmark & \cmark & \xmark & \xmark & 50 & 1.62K \\
miniImageNet~\cite{vinyals2016matching} & image label & \xmark & \cmark & \xmark & \cmark & \xmark & \xmark & 100 & 60K \\
Meta-Dataset~\cite{triantafillou2019meta} & image label & \xmark & \cmark & \cmark & \xmark & \xmark & \xmark & 4,934 & 52.8M \\
ORBIT~\cite{massiceti2021orbit} & frame label & \xmark & \cmark & \cmark & \xmark & \xmark & \xmark & 486 & 2.69M \\
\midrule
RPM~\cite{barrett2018measuring} & shape & \xmark & \xmark & \cmark & \cmark & \xmark & \cmark & 50 & 11.36M \\
V-PROM~\cite{teney2019v} & attributes \& counts & ~~\cmark$^*$ & \cmark & \cmark & \cmark & \xmark & \cmark & 478 & 235K \\
Bongard-LOGO~\cite{nie2020bongard} & shape & \xmark & \xmark & \cmark & \cmark & \xmark & \cmark & 627 & 12K \\
\midrule
Bongard-HOI (ours) & HOI & \cmark & \cmark & \cmark & \cmark & \cmark & \cmark & 242 & 53K \\
\bottomrule
\end{tabular}
\vskip -0.2in
\end{table*}

\vspace{-5pt}
\section{Bongard-HOI Benchmark}
\vspace{-3pt}
For a few-shot binary prediction instance in Bonagrd-HOI, it has a set of positive examples $\calP$, a set of negative samples $\calN$, and a query image $I_q$. Images in $\calP$ depict a certain visual concept (\eg, \texttt{ride bicycle} in Fig.~\ref{fig:teaser}), while images in $\calN$ do not. In each task, there are only six images in both $\calP$ and $\calN$. As a result, a human tester or machine learning model needs to induce the underlying concept from just a few examples. Given the query image $I_q$, a binary prediction needs to be made: whether the certain visual concept depicted in $\calP$ is available in $I_q$ or not. Later, we will detail how to construct these few-shot instances.

\vspace{-5pt}
\subsection{Constructing Bongard Problems}
\vspace{-3pt}
\label{sub_sec:bp_construction}
Few-shot instances in Bongard-HOI are constructed with natural images. We choose to use visual relationships as underlying visual concepts. In our early experiments, we also studied visual attributes to construct few-shot instances, for example, color and shape of bird parts~\cite{WahCUB_200_2011}, facial attributes~\cite{liu2015faceattributes}. But such visual attributes annotations either require too much domain knowledge for human annotators or are too noisy to curate. Another option we investigated is scene graph~\cite{krishna2017visual}, which is a combination of both visual relationships and visual attributes. However, there could be too many convoluted visual concepts in a single image, resulting in ambiguous few-shot instances.

In this paper, we construct few-shot instances on top of the HAKE dataset~\cite{li2019hake,li2020detailed} focusing on human-object interactions. It provides unified annotations following the annotation protocol in HICO~\cite{chao2015hico} for a set of datasets widely used for HOI detection, including HICO~\cite{chao2015hico}, V-COCO~\cite{gupta2015visual}, OpenImages~\cite{OpenImages}, HCVRD~\cite{zhuang2018hcvrd}, and PIC~\cite{liao2020ppdm}. HAKE has 80 object categories, which are consistent with the vocabulary defined in the standard COCO dataset~\cite{lin2014microsoft}. 
It also has 117 action labels, leading to 600 human-object relationships\footnote{Some combinations of objects and actions are infeasible.}. 

Denote a concept $c=\langle s, a, o\rangle$ as a visual relationship triplet, where $s, a, o$ are the class labels of subject, action, and object, respectively. In this paper, $s$ is always \texttt{person}. We start with selecting a set of positive images $\mathcal{I}_c = \{I_1, \dots\}$ from HAKE that depict such a relationship. We also need negative images, where the visual concept $c$ is not contained by them. In specific, we collect another set of images $\mathcal{I}_{\bar{c}}$ with concept $\bar{c}=\langle s, \bar{a}, o\rangle$, where $\bar{a}\neq a$, meaning that we select \emph{hard negatives}. As a result, images from both $\mathcal{I}_{c}$ and $\mathcal{I}_{\bar{c}}$ contain the same categories of objects and the only differences are the action labels, \textit{making it impossible to trivially distinguish positive images from the negatives by doing visual recognition of object categories only}. Rather, detailed visual reasoning about the interactions of human and objects are desired. Fig.~\ref{fig:hard_negatives} illustrates the difficulties introduced by the hard negatives. Finally, as an entire image may contain multiple HOI instances, we use image regions (crops) around each HOI instance instead of the original image to ensure only a single HOI instance is presented in a single image.

Next, we need to sample few-shot instances from the positive images $\mathcal{I}_{c}$ and the negatives $\mathcal{I}_{\bar{c}}$. We randomly sample images to form $\calP$, $\calN$, and a query image $I_q$. Two parameters control the sampling process: $M$, the number of images in $\calP$ and $\calN$ ($M=6$ in Bongard-HOI), and the overlap threshold $\tau$, indicating the maximum number of overlapped images between two few-shot instances. We want to sample as many few-shot instances as possible, but we also need to avoid significant image overlap between few-shot instances, which limits the diversity of the data. We end up setting $\tau=3$ and $\tau=2$ for training and test sets, respectively. More details can be found in the supplementary material.

\vspace{-5pt}
\subsection{Data Curation}
\vspace{-3pt}
Although the HAKE dataset~\cite{li2019hake,li2020detailed} has provided high-quality annotations, we found that curations are still needed to construct few-shot instances. Recall, to sample negative images, we assume a particular action is not depicted in them. In HAKE, an image region may have multiple action labels. Naively relying on the provided annotations is problematic as the action labels are either not manually exclusive or not exhaustively annotated. 
We hire high-quality testers on the Amazon Mechanical Turk (MTurk) platform, who maintain a good job approval record, to curate existing HOI annotations. We discuss the data curation process in detail and show visual examples in detail in the supplementary material.

After the aforementioned data curations, each image region is assigned to a single action label, describing the most salient visual relationship. With the curated annotations, action labels between a person and objects of a certain category are mutually exclusive so that we can significantly reduce the ambiguity when constructing few-shot instances. Finally, we hire high-quality testers on the MTurk platform to further remove the ambiguous few-shot instances in the test set. Every single few-shot instance is assigned to three independent testers. We compare their responses with the ground-truth labels and discard about 2.5\% few-shot instances where none of the three testers correctly classifies the query images. In the end, we report the accuracy of human testers on those left unambiguous few-shot instances as a human study to examine human-level performance on our Bongard-HOI benchmark, where the average accuracy is 91.42\%.

\vspace{-5pt}
\subsection{Generalization Tests}
\vspace{-3pt}

Transferring the knowledge that an agent has seen and learned is a hallmark of visual intelligence, which is a long-stand goal for the entire AI community. It is also a core focus of the Bongard-HOI benchmark. Following~\cite{barrett2018measuring}, we provide multiple test splits to investigate different types of generalization, aiming at a systematic understanding of how the tested models generalize on our benchmark. Specifically, the visual concept we consider in Bongard-HOI is an HOI triplet $\langle s, a, o\rangle$ and we have two variables of freedom: action $a$ and object $o$. Therefore, by controlling whether an action or object is seen during training, we can study generalization to unseen actions, unseen objects, or a combination of two. We end up introducing four separate test sets, as shown in Fig.~\ref{fig:generalization}. We provide detailed statistics on our training and test sets in the supplementary material.

Ideally, after learning from examples of \texttt{sit\_on bed}, a machine learning model can quickly grasp the concept \texttt{sit\_on bench}. More importantly, such a model should learn \emph{how to learn} from just a few examples, so that they can still induce the correct concept (visual relationship) in the most challenging cases, where both actions and objects are not seen during training (\eg, \texttt{shear sheep}).

\vspace{-5pt}
\section{Possible Models for Bongard-HOI}
\label{sec:models}
\vspace{-3pt}

There are many possible ways of tackling Bongard-HOI, such as few-shot learning, conventional HOI detection, etc. We are particularly interested in investigating few-shot learning methods, as our benchmark requires the learner to identify the visual concept with very few samples (positive and negative images in $\calP$ and $\calN$, respectively). To further improve the few-shot learning methods, we consider encoding the images with Relation Network~\cite{santoro2017a}, aiming at better compositionality in the learned representations. Finally, we introduce an oracle model to testify whether Bongard-HOI can be trivially solved using state-of-the-art HOI detection models.  

\vspace{-5pt}
\subsection{Few-shot Learning in Bongard-HOI}\label{sec:few_shot_bongard}
\vspace{-3pt}

We start with a formal definition of the few-shot learning problem in Bongard-HOI. Specifically, each task includes multiple few-shot \emph{instance} with $N=2$ classes and $2M$ samples, \ie, the model learns from a training set $\calS = \calP \cup \calN = \{(I_1^P, 1), \dots, (I_M^P, 1), (I_1^N, 0), \dots, (I_M^N, 0)\}$ and is evaluated on a query image $(I_q, y_q)$. Each example $(I, y)$ includes an image $I \in \mathbb{R}^{H\times W\times 3}$ and a class label $y \in \{0, 1\}$, indicating whether $I$ contains the visual concepts depicted in $\calP$. In Bonagrd-HOI, we set $M=6$ as our default parameter and therefore each few-shot instance is ``2-way, 6-shot''. Following~\cite{triantafillou2019meta}, we propose to solve these few-shot prediction instances with the following two families of approaches:
\vskip 0.05in
\noindent\textbf{Non-episodic methods.} In these methods, a simple classifier is trained to map all the images in a few-shot instance (including images in $\calP$, $\calN$, and the query image) to the class of the query. The classifier can be parameterized as a neural network over some learned image embeddings, \emph{i.e.} representations produced by convolutional neural networks (CNNs). In other words, we view each few-shot instance as a single training sample $(\bigcup_{i=1}^{2M+1}I_i, y_q)$ rather than a few-shot instance with multiple training samples $(I, y)$. Our experiments cover two different ways to encode the images: CNN and Wide Relational Network (WReN)~\cite{barrett2018measuring,nie2020bongard}.
\vskip 0.05in
\noindent\textbf{Meta-learning methods.} These methods adopt the episodic learning setting, \ie, they learn to train a classifier using $2M$ samples from $\calS$ and evaluate their trained classifier on the query $(I_q, y_q)$. In general, their objective (also called \emph{meta-objective}) is to minimize the prediction error on the query. Different meta-learning methods have their own ways to build the classifier and optimize the meta-objective. In our experiments, we consider the following state-of-the-art methods: 1) \textit{ProtoNet} \cite{snell17protonet}, a metric-based method; 2) \textit{MetaOptNet} \cite{lee2019meta} and \textit{ANIL} \cite{raghu2019rapid}, two optimization-based approaches. Moreover, we also use a strong baseline meta-learning model, \textit{Meta-Baseline}~\cite{chen2020new}, which reports competitive results in many few-shot prediction tasks. We refer readers to the related papers for more details.

\vspace{-5pt}
\subsubsection{Image Encoding with Relational Network}
\label{sub_sec:rn}
\vspace{-3pt}
As mentioned above, representation learning of the input images can be crucial to the success of few-shot learning methods on Bongard-HOI. As our benchmark demands learning compostional concepts (HOIs), simply feeding an image into a Convolutional Neural Network (CNN) is not optimal. To this end, we propose to use the Relational Network~\cite{santoro2017a}, which shows promising compositional reasoning accuracy on a Visual Question Answering (VQA) benchmark~\cite{johnson2017clevr}, to explicitly encode the compositionality of visual relationships. In specific, the feature representations of the image $I$ is computed as
\begin{align*}
    \mathrm{RN}(I) = f_{\phi} \circ \sum_{i,j} g_{\psi} \left( \texttt{concat}(h_{\theta}(o_i, I), h_{\theta}(o_j, I))\right),
\end{align*}
where $o_i$ and $o_j$ are two detected objects of the image $I$, provided by ground truth object annotations or a pre-trained object detector like Faster R-CNN~\cite{ren2017faster}. $h_{\theta}$ denotes the RoI Pooled features of $o_i$ from a ResNet backbone~\cite{he2016deep} followed by a MLP (multi-layer perceptron)~\cite{ren2017faster}, which is parameterized by $\theta$. $g_{\psi}$ and $f_{\phi}$ are two additional MLPs.

\begin{figure}
\centering
\renewcommand{\tabcolsep}{1.0pt}
\begin{tabular}{cc}
\includegraphics[height=0.345\linewidth]{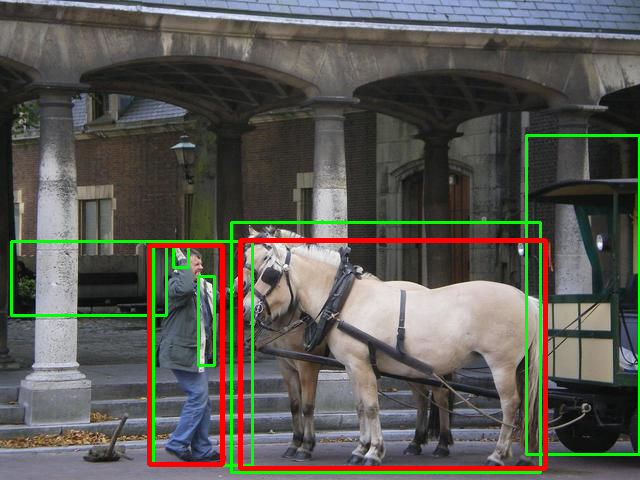} &
\includegraphics[height=0.345\linewidth]{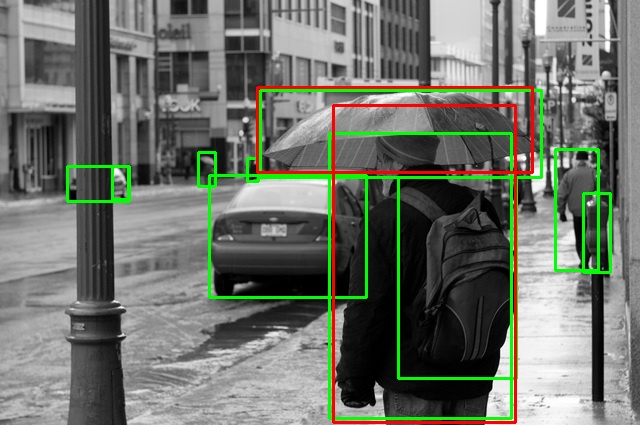}\\ 
\end{tabular}    
\vskip -0.15in
\caption{\textbf{Class-agnostic (objectness) detections.} We show the detections from our class-agonostic detector (in green) and ground-truth human and object boxes (in red). }
\label{fig:objectness_detection}
\vskip -0.25in
\end{figure}

A challenge we are facing is the unseen object categories in the test sets. Since the object detector has to be pre-trained on a dataset without the unseen object categories, it is likely to fail on our test set where images could contain objects belonging to these categories. To tackle this issue, we train a binary class-agnostic (objectness) detection model instead to get $o_i$ and $o_j$. Class-agnostic object detections are shown in Fig.~\ref{fig:objectness_detection}. As we can see, all objects of interest have been successfully detected. But at the same time, there are a lot of other distracting ones, such as the bench and the wagon in the left image of Fig.~\ref{fig:objectness_detection}. This is a unique challenge of dealing with visual reasoning over real-world images. We devote discussions to it in the experiment section.

\vspace{-5pt}
\subsection{Oracle}
\begin{figure}[t]
\centering
\renewcommand{\tabcolsep}{1.5pt}
\newcommand{\loadFig}[1]{\includegraphics[height=0.241\linewidth]{#1}}
\begin{tabular}{cc|cc}
\multicolumn{2}{c|}{$\calP$} & \multicolumn{2}{c}{$\calN$} \\
\loadFig{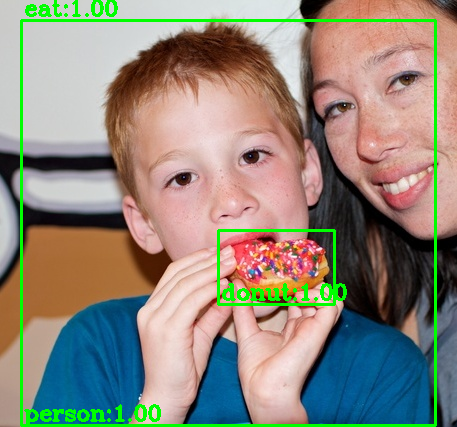} & 
\loadFig{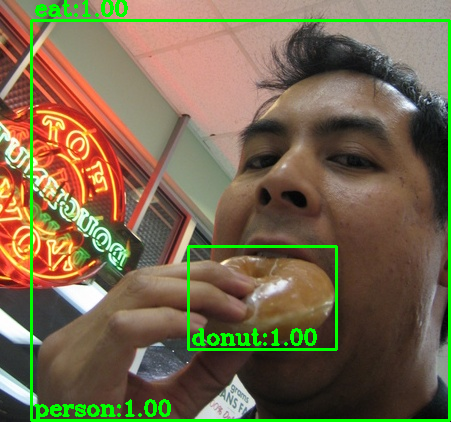} & 
\loadFig{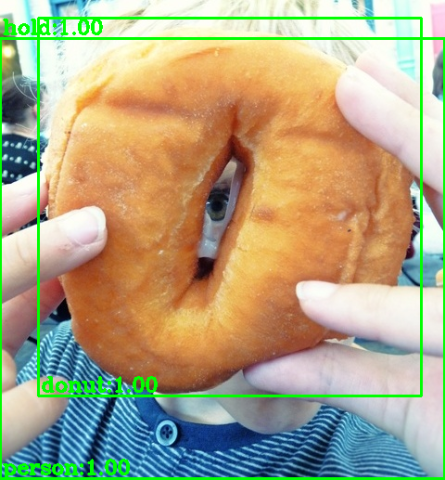} & 
\loadFig{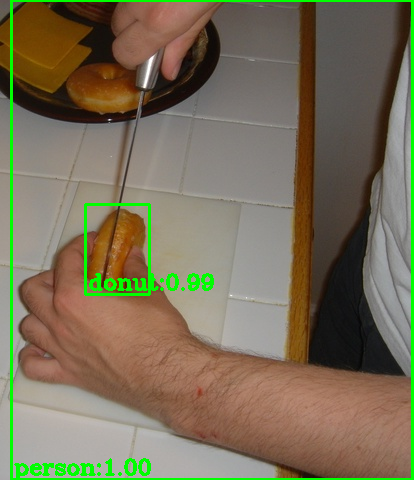} \\
\loadFig{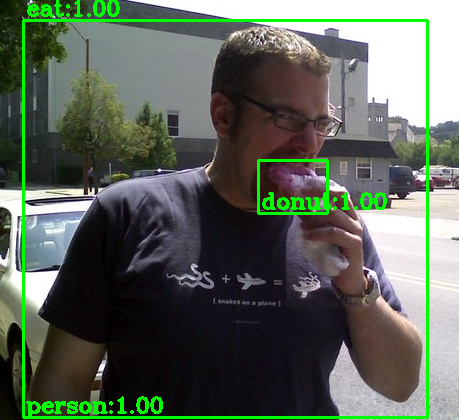} & 
\loadFig{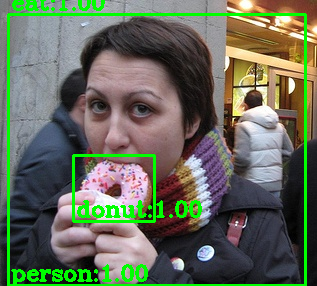} & 
\loadFig{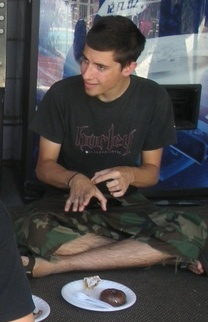} & 
\loadFig{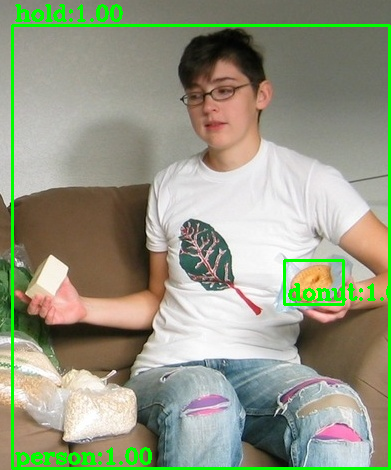} \\ 
\midrule
\multicolumn{4}{c}{Query images:} \\
\multicolumn{2}{r}{\loadFig{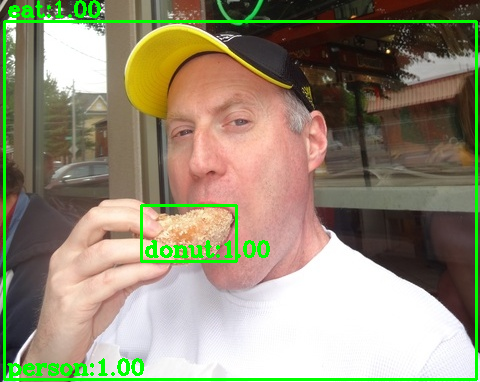}} &
\multicolumn{2}{l}{\loadFig{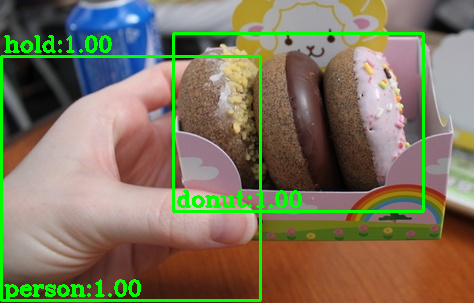}} \\
Predictions: & \multicolumn{1}{c}{\textbf{positive}} & \multicolumn{2}{c}{\textbf{negative}} \\
\end{tabular}
\vskip -0.15in
\caption{\textbf{Illustration of our oracle model.} We first generate some detections for all the images using HOITrans~\cite{zou2021_hoitrans}. Note that some images may not have any detection at all. According to the detections in the $\calP$ and $\calN$, the common concept is \texttt{eat donut}. As a result, in the bottom row, the first query image is considered to be positive as its HOI detections contain \texttt{eat donat}. The second query image is negative. Zoom in for the best view.
}
\label{fig:oracle}
\vskip -0.25in
\end{figure}

\vspace{-3pt}
One may wonder if our Bongard-HOI benchmark could be trivially solved using the state-of-the-art HOI detection model. To address this concern, we develop an oracle model resorting to the HOITrans~\cite{zou2021_hoitrans}, which is based on the Transformer model~\cite{vaswani2017attention} and reports state-of-the-art accuracy on the HICO~\cite{chao2015hico} and V-COCO~\cite{gupta2015visual} benchmarks. In specific, let's denote the HOI detections in the $\calP$ and $\calN$ as $\calD^P$ and $\calD^N$, respectively. $\calD^P$ contains the detections from all of the images in the $\calP$, defined as $\calD^P=\{c_i^P\}_{i=1}^{N_P}$, where $c_i^P$ is a HOI triplet introduced in Section~\ref{sub_sec:bp_construction}. $N_P$ is the total number of detections. Note that there may be multiple or no detections for a single image. Similarly, $\calD^N$ is defined as $\calD^N=\{c_i^N\}_{i=1}^{N_N}$. According to the property of Bongard-HOI, the visual concept $c_P$ should only appear in the $\calP$, not in the $\calN$. We, therefore, compute $c_P$ as
\begin{align*}
    c_P = \texttt{majority\_vote}(\calD^P - \calD^N),
\end{align*}
where $-$ is the set operator for set subtraction. Here we first exclude the HOIs detected in $\calN$ from $\calD^P$, then the majority of the remaining HOIs will be viewed as the visual concept $c_P$. 
Given the detections $\calD^q=\{c_i^q\}_{i=1}^{N_q}$ for the query image $I_q$, our prediction $y$ becomes
\begin{align*}
  y=\left\{\begin{array}{cc}
      1, & \textrm{if}~c_P \in \calD^q, \\
      0, & \textrm{otherwise.}
  \end{array} \right.
\end{align*}
\vskip -0.1in
Discussions of how to deal with the corner cases, \eg, $\texttt{majority\_vote}$ returns more than 1 concept, $\calD^q$ is empty, etc, are provided in the supplementary material. We illustrate how this model works in Fig.~\ref{fig:oracle}, where we show HOI detections in each image.

We call it our oracle model as it has privileged information, \ie, the entire HOI action \& object vocabulary, including those held-out ones in the test set. As we shall we in Section~\ref{sec:experiments}, such an oracle model still struggles on our Bongard-HOI benchmark, achieving only 62.46\% accuracy on average, which is far below the human-level performance of 91.42\%. It suggests that our Bongard-HOI benchmark is not trivial to solve.

\vspace{-5pt}
\section{Experiments}
\label{sec:experiments}
\vspace{-3pt}
\subsection{Implementation Details}
\vspace{-3pt}
We benchmark the models introduced in Section~\ref{sec:models} on Bongard-HOI to test their performance on human-level few-shot visual reasoning. We use a ResNet50~\cite{he2016deep} as an encoder for the input images. We consider different pre-training strategies: 1) no pre-training at all (scratch), 2) pre-trained on the ImageNet dataset with manual labels~\cite{deng2009imagenet}, and 3) latest self-supervised approach~\cite{chen2020improved} pre-trained on ImageNet but without manual labels. We train an Faster R-CNN~\cite{ren2017faster} class-agnostic objectness detection model on the COCO dataset~\cite{patterson2016coco} using a ResNet101~\cite{he2016deep} pre-trained on ImageNet~\cite{deng2009imagenet} as the backbone. We use the RoIPool operation~\cite{ren2017faster} to get feature representations for each bounding box. We also use ground-truth bounding boxes provided in HAKE~\cite{li2019hake} as input to diagnose the effectiveness of the visual perception. In addition to RoIPooled region features, we also concatenate each bounding box's normalized coordinates (center and spatial dimensions) as spatial information to the Relational Network encoder introduced in Section~\ref{sub_sec:rn}.

\definecolor{mygray}{gray}{0.5}
\newcommand{\gtres}[1]{\textcolor{mygray}{#1}}
\begin{table*}[t!]
    \centering
    \caption{\textbf{Quantitative results on the Bongard-HOI benchmark.} All the models use a ResNet50 as the image encoder. For the input of bounding boxes (bbox), we consider two options: from an object detection model (det) and ground-truth annotations (gt). For the ResNet50 encoder, we experiment with different pre-training strategies: no pre-training at all (scratch), pre-trained on the ImageNet dataset with manual labels (IN), and state-of-the-art self-supervised approach MoCoV2~\cite{chen2020improved}. 
    (* denotes that we are unable to get meaningful results; $^\#$ indicates that the trained model has a run-time error during the inference stage since the condition of the QP solver can not be satisfied).}
    \vskip -0.1in
    \label{tab:quantitative}
    \begin{tabular}{l|c|c|c|c|c|c|c}
    \toprule
    & \multirow{3}{*}{bbox} & \multirow{3}{*}{pre-train} & \multicolumn{4}{c|}{test set} & \multirow{3}{*}{avg.} \\
    \cmidrule{4-7}
    &  & & seen act., & seen act., & unseen act., & unseen act., &  \\
    & & & seen obj. & unseen obj. & seen obj. & unseen obj. \\
    \midrule
    CNN-Baseline~\cite{nie2020bongard} & - & scratch & 50.03 & 49.89 & 49.77 & 50.01 & 49.92 \\
    
    WReN-BP \cite{barrett2018measuring,nie2020bongard} & - & IN & 50.31 & 49.72 & 49.97 & 49.01 & 49.75 \\
    
    \midrule

    ProtoNet* \cite{snell17protonet} & det & IN & - & - & - & - & - \\

    \gtres{ProtoNet} \cite{snell17protonet} & \gtres{gt} & \gtres{IN} & \gtres{58.90} & \gtres{58.77} & \gtres{57.11} & \gtres{58.34} & \gtres{58.28} \\

     \midrule
     
    MetaOptNet$^\#$ \cite{lee2019meta} & det & IN & - & - & - & - & - \\

    \gtres{MetaOptNet} \cite{lee2019meta} & \gtres{gt} & \gtres{IN} & \gtres{58.60} & \gtres{58.28} & \gtres{58.39} & \gtres{56.59} & \gtres{57.97} \\

     \midrule
     
    ANIL \cite{raghu2019rapid} & det & IN & 50.18 & 50.13 & 49.81 & 48.83 & 49.74 \\

     \gtres{ANIL} \cite{raghu2019rapid} & \gtres{gt} & \gtres{IN} & \gtres{52.73} & \gtres{50.11} & \gtres{49.55} & \gtres{48.19} & \gtres{50.15} \\

     \midrule
     
    Meta-Baseline \cite{chen2020new} & det & scratch & 54.61 & 53.79 & 54.58 & 53.94 & 54.23 \\
    
    Meta-Baseline \cite{chen2020new} & det & MoCoV2 & 55.23 & 54.54 & 54.32 & 53.11 & 54.30 \\
        
    Meta-Baseline \cite{chen2020new} & det & IN & 56.45 & 56.02 & 55.60 & 55.21 & 55.82 \\

    \gtres{Meta-Baseline} \cite{chen2020new} & \gtres{gt} & \gtres{IN} & \gtres{58.82} & \gtres{58.75} & \gtres{58.56} & \gtres{57.04} & \gtres{58.30} \\

    \midrule
    HOITrans \cite{zou2021_hoitrans} (oracle) & - & - & 59.50 & 64.38 & 63.10 & 62.87 & 62.46 \\
    \midrule
     Human (Amateur) & - & - & 87.21 & 90.01 & 93.61 & 94.85 & 91.42 \\
    \bottomrule
    \end{tabular}
    \vskip -0.2in
\end{table*}

\vspace{-5pt}
\subsection{Quantitative Results}\label{sec:exp_analysis}
\vspace{-3pt}

The quantitative results of different models on our Bongard-HOI benchmark can be found in Table~\ref{tab:quantitative}. We make the following observations:
First of all, despite the overall difficulties brought by our benchmarks, most models perform worse on the challenging test splits, where actions and/or object categories are completely unseen during training. This observation aligns well with our hypothesis, \emph{i.e.} existing machine learning approaches can be limited in terms of generalizing beyond training concepts. It also echos the findings in Bongard-LOGO~\cite{nie2020bongard}, a dataset studying a similar problem with synthetic images.
Second, meta-learning approaches generally tend to perform better than non-episodic counterparts, which can be on par with or even worse than random guesses (50\% chance). We hypothesize the reason to be the focus on \emph{learning to learn} in these methods, which is essentially required to solve the few-shot instances in the Bongard-HOI benchmark, especially for the challenging test splits with novel categories. Similar observations have also been made in Bongard-LOGO. Moreover, some meta-learning models are distracted by bounding boxes provided by an object detection model. We will discuss this issue in the next section.

Surprisingly, the oracle model (\emph{HOITrans}) also struggles on our tests with an averaged accuracy of 62.46\%, albeit being trained with direct HOI supervision and all action\&object categories. It suggests a clear gap between the existing HOI detection datasets, \eg HAKE~\cite{li2019hake} and Bongard-HOI, where the latter one requires capabilities beyond perception, \eg HOI recognition. Rather, a model might also need context-dependent reasoning, learning-to-learn from very few examples, etc., to perform well on our benchmarks.

Finally, machine learning models still largely fall behind amateur human testers (\eg, 55.82\% of Meta-Baseline vs 91.42\%).  While we only give human testers a couple of examples about visual relationships before they start working on solving Bongard-HOI, they can quickly learn how to induce visual relationships from just a few examples, reporting an average 91.42\% accuracy on our Bongard-HOI benchmark. Particularly, there are no significant differences for the different subsets of the test set. We hope our findings will foster more research efforts on closing this gap. 

\vspace{-5pt}
\subsection{Discussions}
\vspace{-3pt}

\noindent\textbf{We need holistic perception and reasoning.} Our work suggests that the significant challenges in current visual reasoning systems lie in both the reliability of perception and the intricacy of the reasoning task itself. Models that have only good pattern recognition performances are likely to fail on our benchmarks. Rather, an ideal learner needs to integrate visual perception in natural scenes and detailed cognitive reasoning as a whole. This marks our key motivation to propose Bongard-HOI as the first step towards studying these two problems holistically.

\noindent\textbf{Pre-training improves performances.} Intuitively, models for Bongard-HOI might need additional representation learning, \eg pre-training, since currently we only train on binary labels of few-shot instances. We can see from Table~\ref{tab:quantitative} that \emph{pre-training is very helpful}. Compared to no pre-training, using either manual labels or self-supervision leads to a performance boost. In particular, the self-supervised pre-training~\cite{chen2020improved} does not use any manual labels for supervision. Yet it can produce better results than learning from scratch.

\noindent\textbf{Visual perception matters in Bongard-HOI.} Finally, an imperfect perception could still be a major obstacle here. 
Different from Bongard-LOGO~\cite{nie2020bongard} which uses synthetic shapes instead, Bongard-HOI studies visual reasoning on natural scenes, which often contain rich visual stimuli, issuing such as large intra-class variance and cluttered background also present challenges to reliable visual perception on which reasoning is based.
In our case, bounding boxes produced by an object detection model can be inevitably noisy. 
Some meta-learning models, including ProtoNet~\cite{snell17protonet}, have difficulties inducing the true visual relationships. For MetaOptNet~\cite{lee2019meta}, although we can finish training, we constantly encounter run-time errors where the condition of the QP solver is not satisfied during the inference stage. Instead, when taking clean ground-truth bounding boxes as input, all of these approaches produce better accuracy. Note that using ground-truth bounding boxes only serves as an oracle, which does not indicate the models' authentic performance.

\vspace{-5pt}
\section{Related Work}
\vspace{-3pt}

\paragraph{Visual relationship detection benchmarks.} Various benchmarks are also dedicated for visual relationship recognition and detection, particularly for human-centric relationships (\ie, HOI). 
In the seminal work of Visual Genome~\cite{krishna2017visual}, scene graph annotations, including relationships of different objects, are provided. A subset of the annotations is used in VRD~\cite{lu2016visual} to focus on visual relationship detection. In a recent effort, large-scale visual relationships are provided in the Open Images dataset~\cite{OpenImages}. HOI, is of particular interest to understand the interactions of humans and other objects. A lot of HOI benchmarks, such as HICO~\cite{chao2015hico}, COCO-a~\cite{ronchi2015describing}, vCOCO~\cite{gupta2015visual}, and HOI-COCO~\cite{hou2021fcl}, are built on top of the object categories provided in the COCO dataset~\cite{lin2014microsoft}. The MECCANO~\cite{ragusa2021meccano} dataset focuses on human-object interactions in egocentric settings and industrial scenarios. Ambiguous-HOI~\cite{li2020detailed} is part of the HAKE project~\cite{li2019hake}, where the focus is human activity understanding with a large-scale knowledge base and visual reasoning.

Although our Bongard-HOI benchmark is built on top of the dataset HAKE~\cite{li2019hake}, it differs from the existing visual relationship and HOI benchmarks, since we focus on human-level cognitive reasoning instead of recognition. To solve Bongard-HOI, one might not need to explicitly name the underlying visual relationship but does need to induce the HOI from a few images and perform context-dependent reasoning. Our results also suggest that Bongard-HOI cannot be trivially solved by the state-of-the-art models on these datasets, \eg HOITrans~\cite{zou2021_hoitrans}.

\vskip 0.05in
\noindent\textbf{Few-shot and meta learning models.} Few-shot learning aims at learning from a limited number of training samples ~\cite{fe2003bayesian, koch2015siamese}. With the goal of extracting the generic knowledge across tasks and generalizing to a new task using task-specific information, meta-learning (or learning-to-learn)~\cite{hochreiter2001learning} becomes one of the leading approaches to deal with the few-shot learning problems. In general, meta-learning methods are divided into three categories: 1) memory-based methods, such as MANN \cite{santoro2016meta} and SNAIL \cite{mishra2018simple}, 2) metric-based methods, such as Matching Networks~\cite{vinyals2016matching} and ProtoNet \cite{snell17protonet}, and 3) optimization-based methods, such as MetaOptNet \cite{lee2019meta} and ANIL~\cite{raghu2019rapid}. 

These meta-learning methods have been evaluated on several commonly used few-shot learning benchmarks, including miniImageNet~\cite{vinyals2016matching} and tieredImageNet~\cite{ren2018meta}. Although state-of-the-art meta-learning algorithms have achieved excellent performance on these standard few-shot image classification benchmarks, whether these approaches can generalize to tasks where the concepts to learn (in a few-shot manner) are compositional, \eg visual relationships rather than simple object categories is unknown~\cite{kato2018compositional,hou2020visual}. In other words, existing benchmarks fail to account for the challenging problem of generalizing to new compositional concepts in few-shot learning. Therefore, with a focus on the more challenging visual concepts of visual relationships, we propose Bongard-HOI to serve as a new benchmark for the few-shot learning methods. We believe that our benchmark can foster the development of new few-shot learning, especially meta-learning algorithms to achieve better performances on learning and generalizing with compositional concepts.
\vskip 0.05in
\noindent\textbf{Abstract visual reasoning benchmarks.} Inspired by cognitive studies, several benchmarks have been built for abstract reasoning, highlighting cognitive abstract reasoning. Notable examples include compositional question answering~\cite{johnson2017clevr,ma2022relvit}, physical reasoning~\cite{bakhtin2019phyre,yi2020clevrer}, math problems~\cite{saxton2019analysing}, and general artificial intelligence~\cite{chollet2019measure,xie2021halma}. The most relevant to our benchmark are RPM~\cite{barrett2018measuring}, its variant with natural images~\cite{teney2019v}, and Bongard problems with synthetic shapes~\cite{nie2020bongard} and physical problems~\cite{weitnauer2012physical}. While most of them consider synthetic images~\cite{barrett2018measuring,nie2020bongard,weitnauer2012physical}, our Bongard-HOI benchmark studies cognitive reasoning on natural images, which impose unique challenges due to the difficulty of visual perception. Moreover, we use human-object interaction as the underlying concepts to construct few-shot instances, which require explicit compositional concept learning in a few-shot manner, compared to the object categories and shapes~\cite{teney2019v}. Moreover, the existence of hard negatives in the few-shot instances makes our benchmark more challenging.

\vspace{-5pt}
\section{Conclusion}
\vspace{-3pt}
In this paper, we introduced the Bongard-HOI benchmark focusing on the few-shot learning and the generalization with compositional concepts in real-world visual relationship reasoning. Drawing inspirations from the classic Bongard problems~\cite{bongard1968recognition}, we constructed few-shot instances using the visual relationships between humans and objects as the underlying concepts. Our benchmark is built on top of an existing HOI dataset, HAKE~\cite{li2019hake}, where we carefully curated the provided annotations to construct the few-shot instances. We benchmarked state-of-the-art few-shot learning methods, including both non-episodic and meta-learning approaches. Our findings suggested that current machine learning models still struggle to generalize beyond concepts that they have seen during the training process. Moreover, natural images in our benchmark contain rich stimuli, imposing great challenges to the machine learning models in the real-world visual relationship reasoning tasks. By building the Bongard-HOI benchmark, we hope to foster research efforts in real-world visual relationship reasoning, especially in holistic perception-reasoning systems and better representation learning.

\appendix

\newcommand{\calA}{\mathcal{A}}

\section{Limitation Statement}
We re-use the images collected by the HAKE~\cite{chao2015hico} creators, including the ones for HICO~\cite{chao2015hico}, V-COCO~\cite{gupta2015visual}, OpenImages~\cite{OpenImages}, HCVRD~\cite{zhuang2018hcvrd}, and PIC~\cite{liao2020ppdm}, which were crawled from the web. Except the images, in this paper, no identity related information were collected nor used when constructing the dataset and benchmarking other approaches. It is possible, however, that some person may be identified via facial recognition techniques. We will provide contact information of the benchmark maintainer and commit to processing request of removing some certain images from the dataset. In addition, similar to other human-centric dataset, the images we use are from just a small portion of the population, which may contain biases toward some certain races, gender, ethnic groups, etc.
We are unable to measure the bias as we do not have any identity-related data. We encourage researchers to investigate such issues.

\section{More details on the Bongard-HOI\\Benchmark}

\subsection{Constructing Bongard Problems}
Given positive images $\mathcal{I}_c$ that depict a certain relationship $c=\langle s, a, o\rangle$ and and negative images $\mathcal{I}_{\bar{c}}$ that does not, we need to sample few-shot instances from them. We randomly sample images to form $\calP$, $\calN$, and a query image $I_q$. Two parameters control the sampling process: $M$, the number of images in $\calP$ and $\calN$ ($M=6$ in Bongard-HOI), and the overlap threshold $\tau$, indicating the maximum number of overlapped images between two few-shot instances. We want to sample as many few-shot instances as possible, but we also need to avoid significant image overlap between few-shot instances, which limits the diversity of the data. The sampling process in summarized in Algorithm~\ref{alg:sample_few-shot instances}. We set $\tau=3$ and $\tau=2$ for training and test sets, respectively. 

        \begin{algorithm}[h]
        \KwIn{Positive images $\mathcal{I}_c$, negative images $\mathcal{I}_{\bar{c}}$, number of images in a few-shot instance $M$, overlap threshold $\tau$.}
        \KwOut{Sampled few-shot instances $\calQ$.}
         $\calQ=\emptyset$\;
         \While{True}{
          $\calP^i, \calN^i, I_q^i=sample\_instance(\mathcal{I}_{c}, \mathcal{I}_{\bar{c}}, M)$\;
          \If{\textrm{sample fails}}{
            break\;
          }
          $t = overlap(\calP^i, \calN^i, I_q^i, \calQ)$\;
          \If{$t < \tau$}{
            $\calQ = \calQ \cup (\calP^i, \calN^i, I_q^i)$\;
          }
         }
         \caption{Sample few-shot instances for a visual concept $c$}
         \label{alg:sample_few-shot instances}
        \end{algorithm}

\begin{figure*}[t]
\centering
\renewcommand{\tabcolsep}{1.5pt}
\newcommand{\loadFig}[1]{\includegraphics[height=0.15\linewidth]{#1}}
\begin{tabular}{ccccc}
scratch, pet & 
eat, hold & 
drive &
eat & 
lie\_on \\
\loadFig{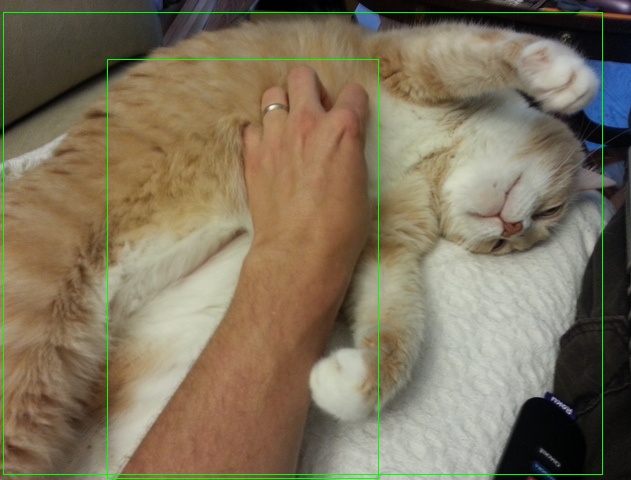} & 
\loadFig{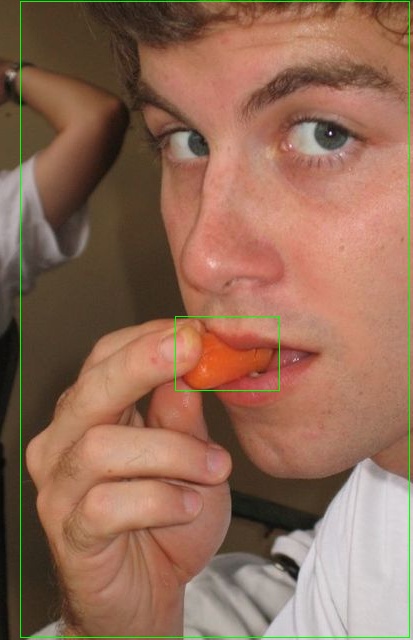} & 
\loadFig{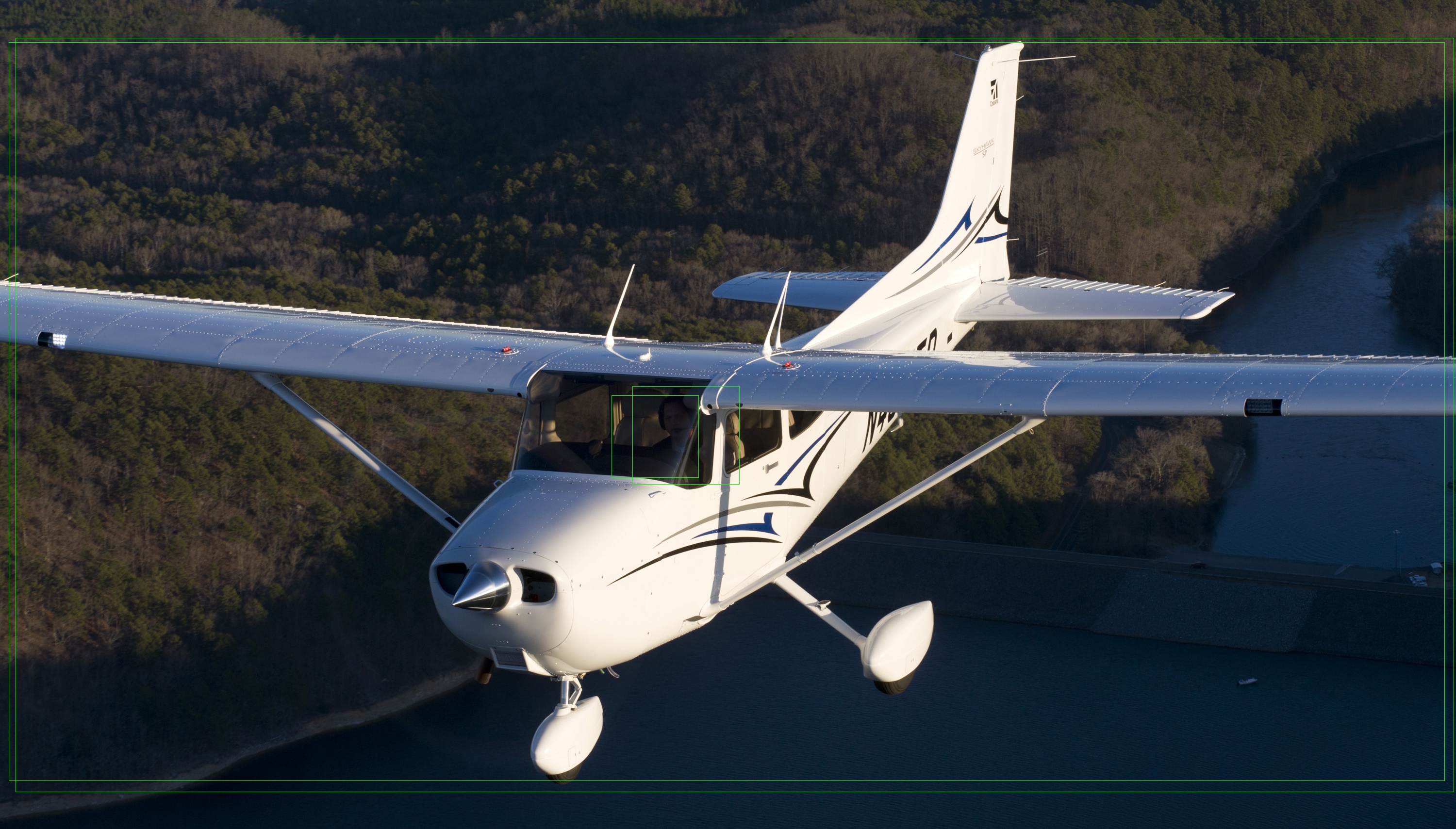} &
\loadFig{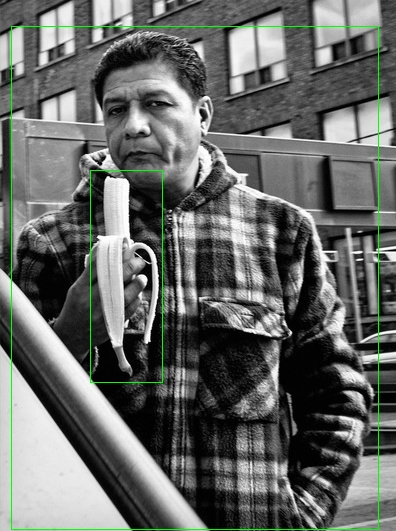} & 
\loadFig{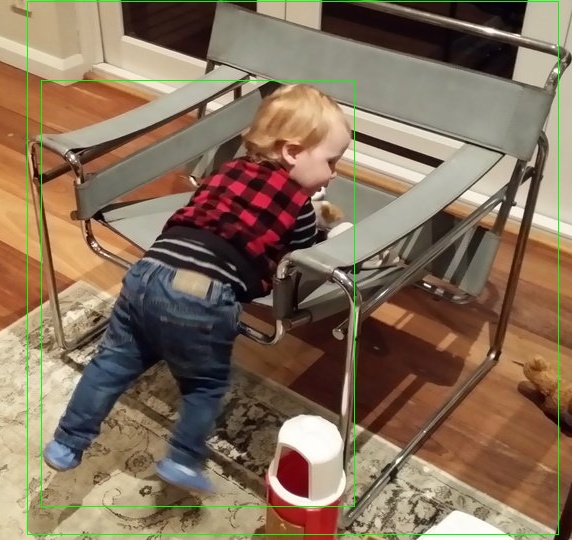} \\

carry, hold & 
jump, ride &
stand\_on & 
check &
inspect \\
\loadFig{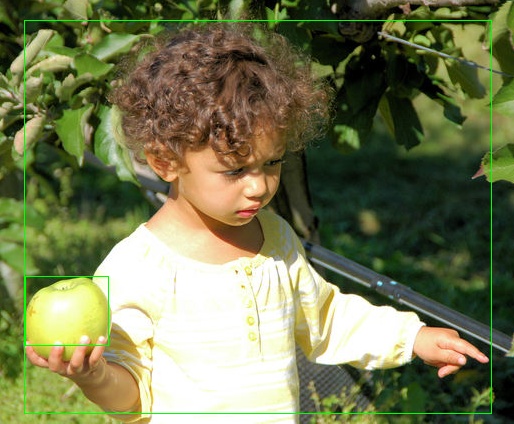} & 
\loadFig{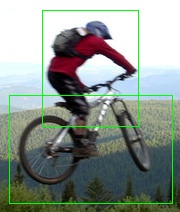} & 
\loadFig{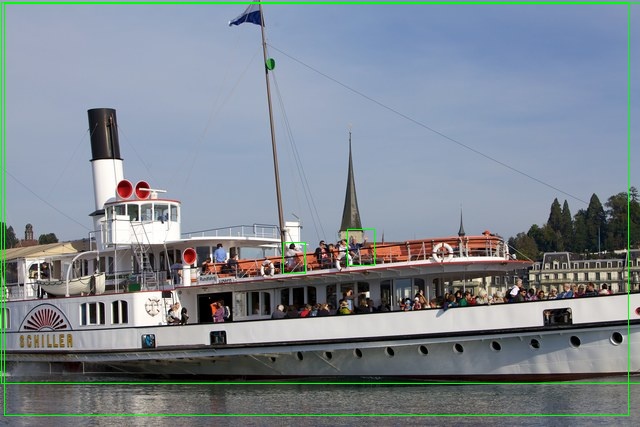} & 
\loadFig{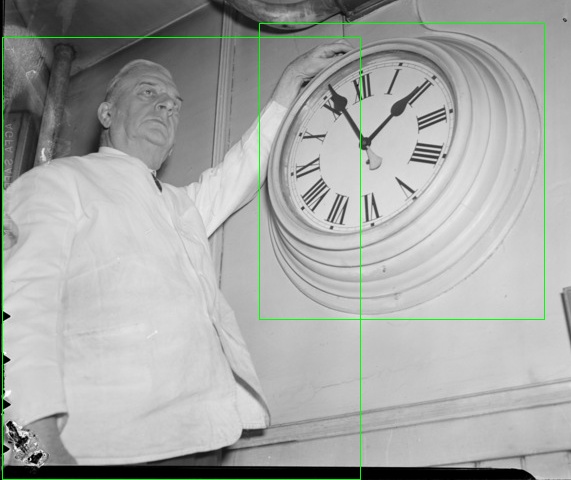} & 
\loadFig{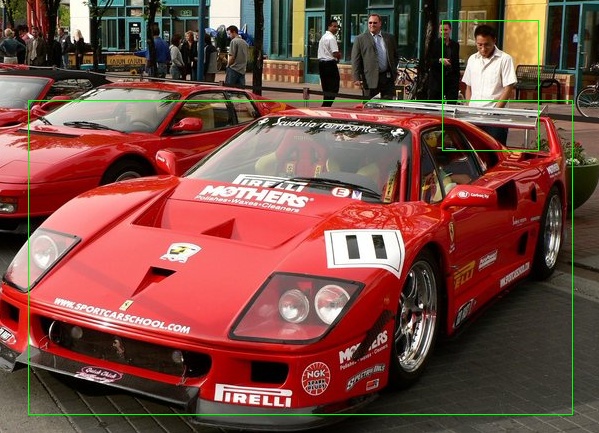} \\

(a)  & (b)  & (c)  & (d)  & (e) \\
\end{tabular}
\caption{\textbf{Samples of annotations where curations are needed.} For each image region, its annotated action labels are show on its top and bounding boxes corresponding to the person and object are shown for visualization purpose. From left to right: (a) similar actions, (b) hierarchical annotations, (c) hard-to-see objects, (d) extrapolating annotations, and (e) inaccurate or confusing annotations.}
\label{fig:curation}
\vspace{-12pt}
\end{figure*}

\subsection{Data Curation}
Although the HAKE dataset~\cite{li2019hake} has provided high-quality annotations, we found that curations are still needed to construct the Bongard problems (few-shot instances) for our Bongard-HOI benchmark. Recall, to sample negative images, we assume a particular action is not depicted in them. In HAKE, an image region may have multiple action labels. Naively relying on the provided annotations is problematic as the action labels are either not manually exclusive or not exhaustively annotated. We show different cases of data curations in Fig.~\ref{fig:curation} and discuss them in details as follows.

\noindent\textbf{Similar actions.} Although some action labels may convey different semantic meanings, for some certain object categories, they look visually similar and indistinguishable. As shown in Fig.~\ref{fig:curation}(a), \texttt{scratch cat} and \texttt{pet cat} are hard to differentiate visually. If we simply use images of \texttt{scratch cat} as negatives to construct few-shot instances for \texttt{pet cat}, such few-shot instances are ambiguous, as it violates the basic assumption that the visual concept depicted in the Set $\calA$ is not available in the Set $\calB$. We therefore simply merge such similar action labels to reduce the visual ambiguity. 

\noindent\textbf{Hierarchical actions.} Action labels are inherently hierarchical. For example, as shown in Fig.~\ref{fig:curation}(b), \texttt{eat carrot} very likely also means \texttt{hold carrot} visually. There are two problems to construct few-shot instances with multiple hierarchical action labels associated with the same image region. First of all, as we previously explained, using images of \texttt{eat carrot} as negatives for \texttt{hold carrot} may cause ambiguity. More importantly, there is the \emph{visual specificity} issue. People tend to focus on capturing the most salient actions in an image, which are usually the parent actions (\texttt{eat carrot} in this case). In our preliminary experiments, images of \texttt{eat carrot} were used as positives for \texttt{hold carrot} to construct few-shot instances. We found that it caused a lot of confusion for human testers. To this end, we merge such hierarchical action labels for the same region, keeping the parent action labels only.

\noindent\textbf{Hard-to-see objects.} In some cases, the person or the objects in image regions are hard to see. For example, in Fig.~\ref{fig:curation}(c), the person with the action label \texttt{stand\_on boat} is hard to see clearly. On the one hand, it causes significant challenges for a visual perception system (\eg,~\cite{he2020mask}) to accurately localize the meaningful objects. At the same time, it also imposes difficulty for annotators to accurately annotate the image region. We simply discard all image regions with hard-to-see objects.

\noindent\textbf{Extrapolating actions.} Actions are continuous. As a result, annotators tend to \emph{extrapolate} the action label given a single image, instead of describing the current state the action. For example, as we can see in the top row of Fig.~\ref{fig:curation}(d), the \texttt{eat} action is about to happen. Yet, the action is different from a normal \texttt{hold banana} without any indication of \texttt{eat}. To distinguish different scenarios, we introduce \texttt{hold\_not\_about\_to\_eat banana}, \texttt{hold\_and\_about\_to\_eat banana}, and \texttt{eat banana}. In this way, all the actions are mutually exclusive. We can sample image regions for form few-shot instances without worrying about causing ambiguity.

\noindent\textbf{Inaccurate or confusing actions.} In some rare cases, the annotations in HAKE are inaccurate or confusing, as shown in Fig.~\ref{fig:curation}(e). We modify the action labels if such a image region depicts a clear action label. Otherwise we discard such regions to avoid introducing ambiguity to sampled few-shot instances.

\noindent\textbf{MTurk data curation.} After performing the aforementioned data curations, each image region is assigned to a single action label, describing the most salient content. Such action labels are mutually exclusive so that we can significantly reduce the ambiguity when constructing few-shot instances. Finally, we hire high-quality testers on the Amazon Mechanical Turk (MTurk) platform, who maintain a good job approval record, to curate the testing set to further remove the ambiguous few-shot instances. Every single BP is assigned to three independent testers. We compare their responses with the ground-truth labels and disard about 2.5\% few-shot instances where none of the three testers correctly classifies the query images. We provide more details of the MTurk curations in Section~\ref{sec:mturk}.

\begin{table}[t]
\small
\begin{subtable}[b]{\linewidth}
\centering
\begin{tabular}{c|c|c}
    \toprule
     & seen object & unseen object \\
    \midrule
    seen action & 99 / 5008  &  36 / 5002 \\
    unseen action  & 20 / 3402 & 12 / 3775 \\ 
    \bottomrule
\end{tabular}
\subcaption{validation set}
\end{subtable}
\vfill
\begin{subtable}[b]{\linewidth}
\centering
\begin{tabular}{c|c|c}
    \toprule
     & seen object & unseen object \\
    \midrule
    seen action & 102 / 4476  &  27 / 4562 \\
    unseen action  & 21 / 3291 & 16 / 1612 \\ 
    \bottomrule
    \end{tabular}
    \subcaption{test set}
\end{subtable}
\caption{\textbf{Number of concepts and few-shot instances in the validation and test sets.} Depending on whether an action and object is seen during the training, we divide the validation and test sets into four categories, where we can study the systematic generalization of machine learning models. For each category, we show number of concepts (combinations of action and object) and number of few-shot instances.}
\label{tab:val_test_stats}
\end{table}

\subsection{Dataset statistics}

\begin{figure*}
\centering
\renewcommand{\tabcolsep}{1.5pt}
\newcommand{\loadFig}[1]{\includegraphics[align=t, height=0.15\linewidth]{#1}}
\begin{tabular}{cc|cc|c}
\multicolumn{5}{c}{Visual concept: \textcolor{red}{\texttt{<person, drink\_with, cup>}}} \\
\multicolumn{2}{c|}{Positive examples} & \multicolumn{2}{c|}{Negative examples} & Query \\
\loadFig{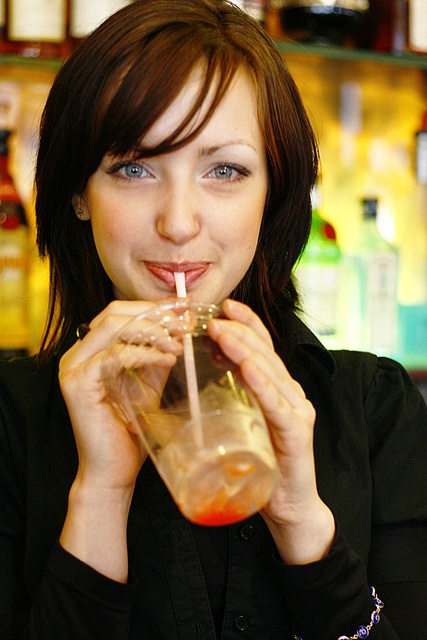} & 
\loadFig{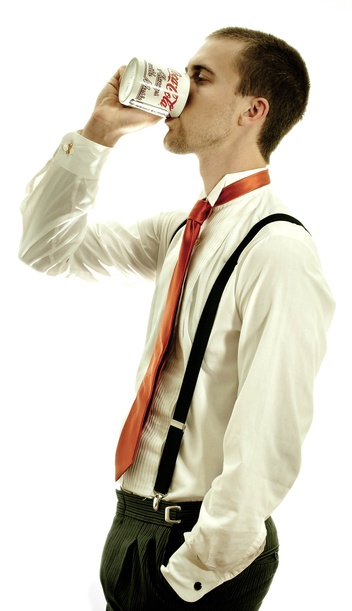} & 
\loadFig{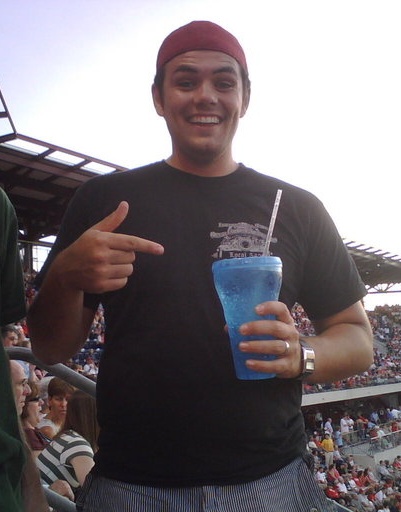} & 
\loadFig{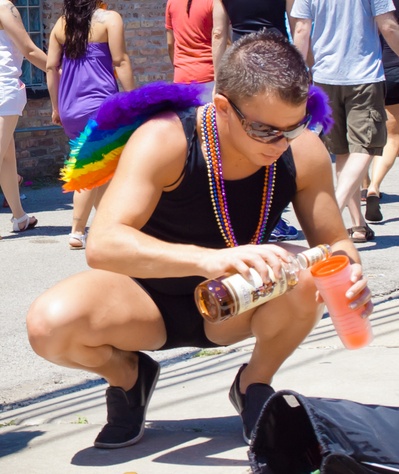} & 
\loadFig{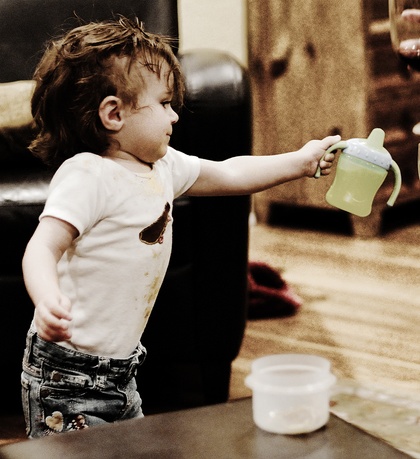} \\
\loadFig{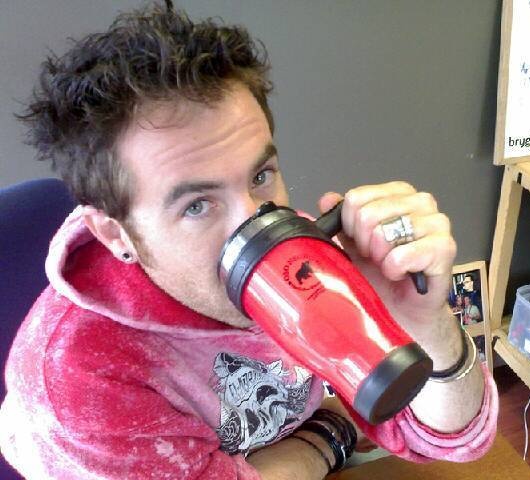} & 
\loadFig{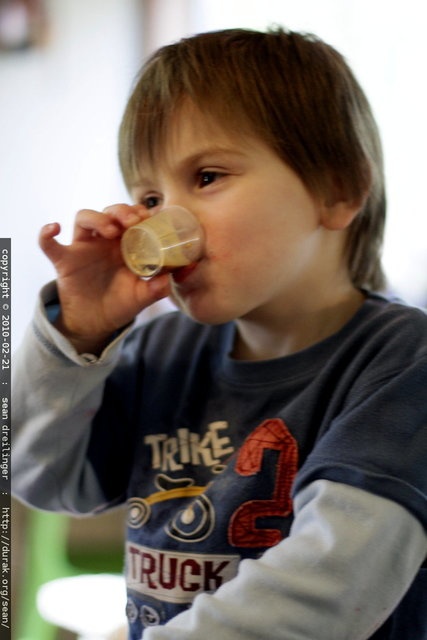} & 
\loadFig{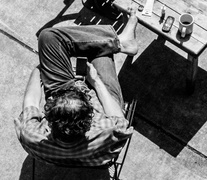} & 
\loadFig{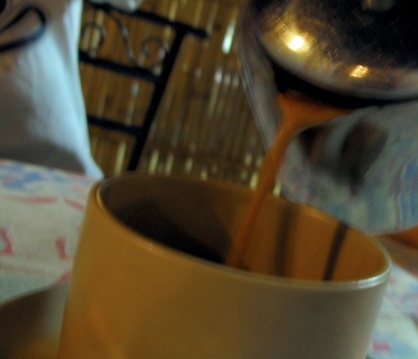} & 
\textbf{label: negative (0)}\\
\loadFig{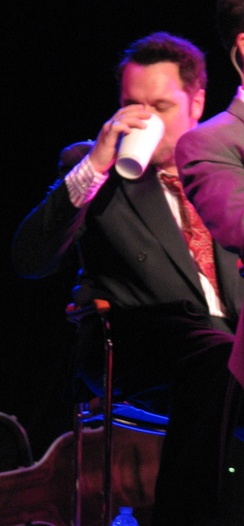} & 
\loadFig{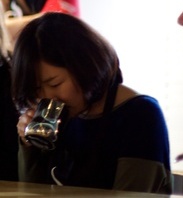} & 
\loadFig{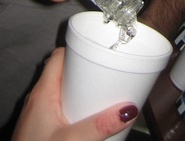} & 
\loadFig{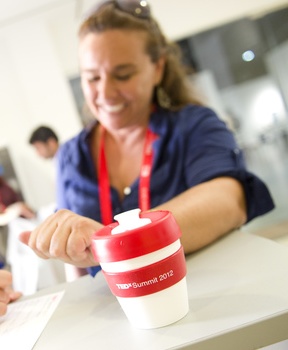} & 
 \\
& & & & \\
\multicolumn{4}{c}{(a)} \\
\multicolumn{5}{c}{Visual concept: \textcolor{red}{\texttt{<person, hold\_but\_not\_drink\_with, cup>}}} \\
\multicolumn{2}{c|}{Positive examples} & \multicolumn{2}{c|}{Negative examples} & Query \\
\loadFig{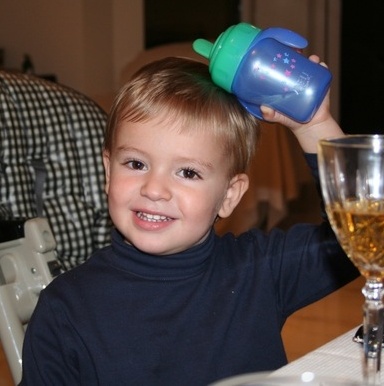} & 
\loadFig{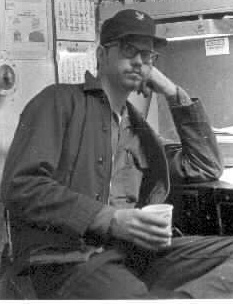} & 
\loadFig{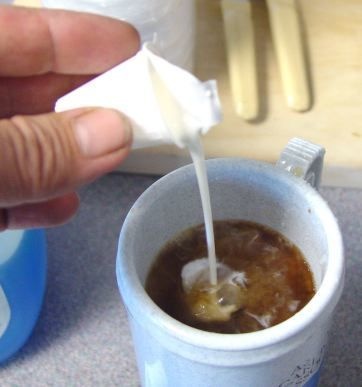} & 
\loadFig{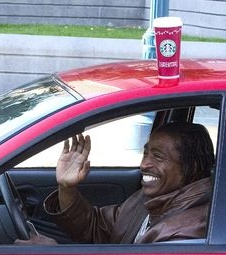} & 
\loadFig{figs/context_dependency/query.jpg} \\
\loadFig{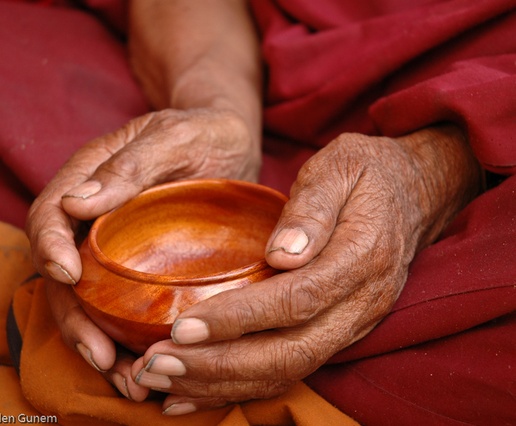} & 
\loadFig{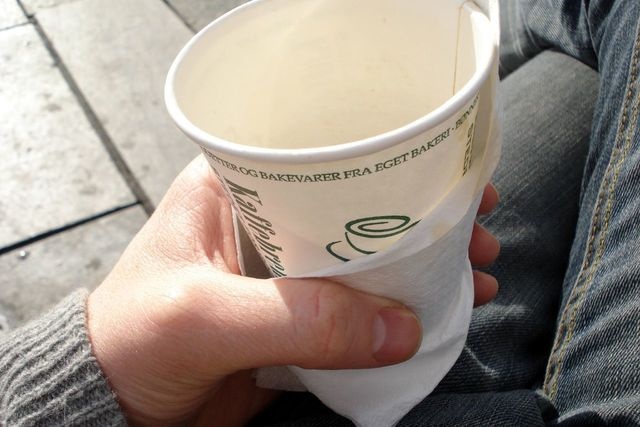} & 
\loadFig{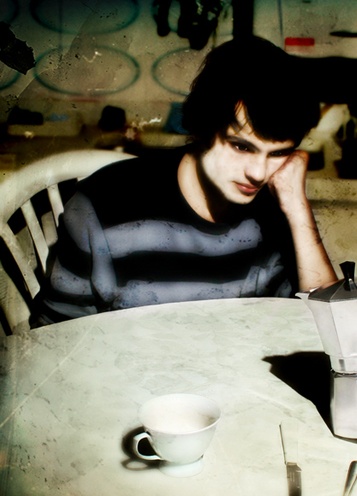} & 
\loadFig{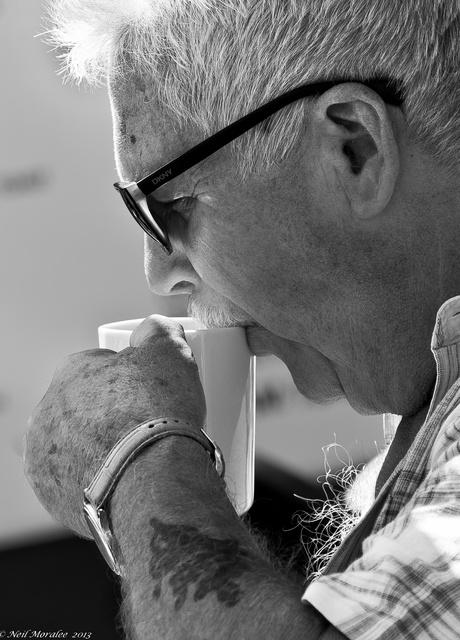} & 
\textbf{label: positive (1)} \\
\loadFig{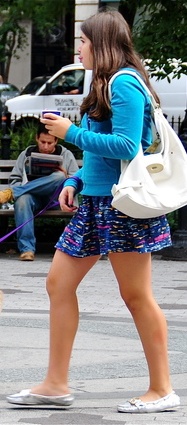} & 
\loadFig{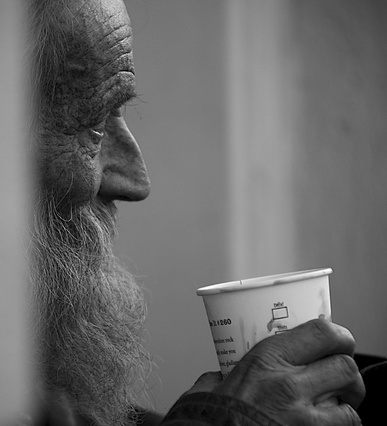} & 
\loadFig{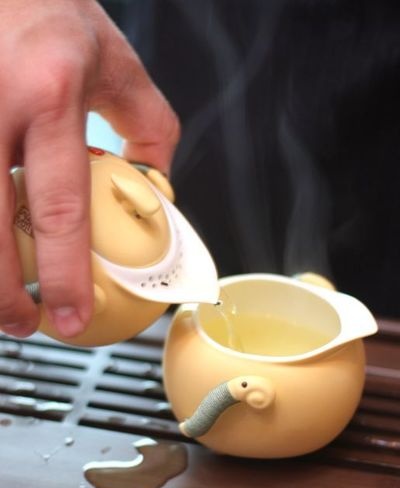} & 
\loadFig{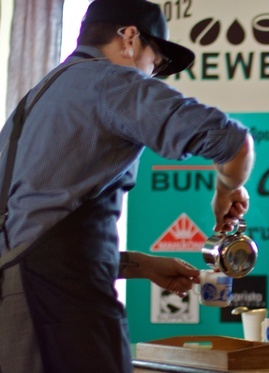} & 
\\
 \multicolumn{4}{c}{(b)}
\end{tabular}
\caption{\textbf{Illustration of the context-dependent reasoning property of the Bongard problems (few-shot instances) in our Bongard-HOI benchmark.} Two instances are shown here with their underlying visual concepts (relationships) displayed on top with red color. The same query image receives two different labels (negative in the top and positive in the bottom) among different context (\ie, positive and negative examples).}
\label{fig:context}
\end{figure*}

Our Bongard-HOI benchmark provides disjoint training, validation, and testing sets.  In specific, there are 118 concepts (visual relationships) and 21,956 few-shot instances in the training set. There are 17,184 and 13,941 few-shot instances in the validation and testing set, respectively, corresponding to 167 and 166 visual concepts. Detailed distribution of concepts and few-shot instances among different generalization types are provided in Table~\ref{tab:val_test_stats}.

\subsection{Illustration about the Context-Dependent Reasoning Property}
Two Bongard problems (few-shot instancess) are shown in Fig.~\ref{fig:context}. For the same query image, among different context (\ie, positive and negative examples), it receives different classification labels. This context-dependent reasoning property distinguishes our Bongard-HOI benchmark from other few-shot learning ones, where an image always has a fixed label.

\begin{figure*}[t]
\centering
\renewcommand{\tabcolsep}{1.5pt}
\newcommand{\loadFig}[1]{\includegraphics[height=0.121\linewidth]{#1}}
\begin{tabular}{cc|cc}
\multicolumn{2}{c|}{$\calP$} & \multicolumn{2}{c}{$\calN$} \\
\loadFig{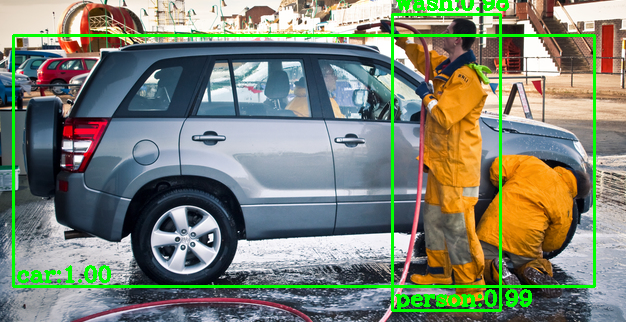} & 
\loadFig{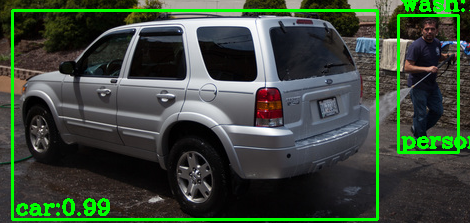} & 
\loadFig{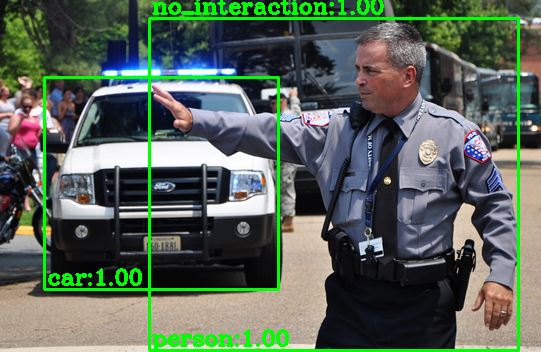} & 
\loadFig{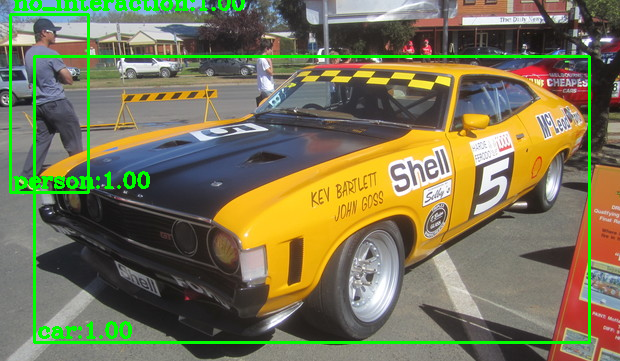} \\
\loadFig{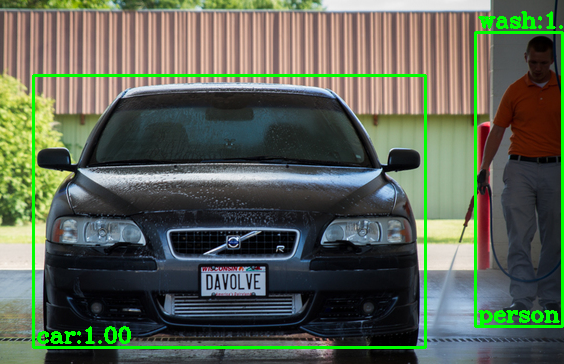} & 
\loadFig{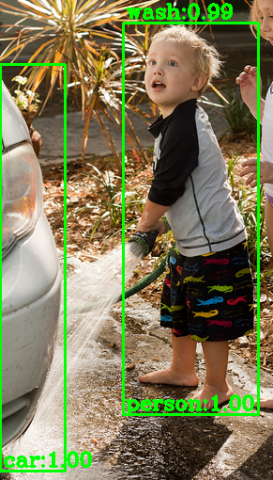} & 
\loadFig{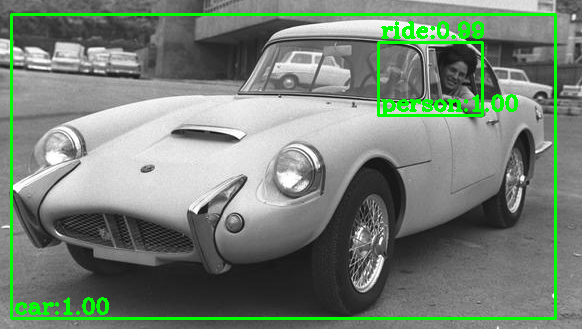} & 
\loadFig{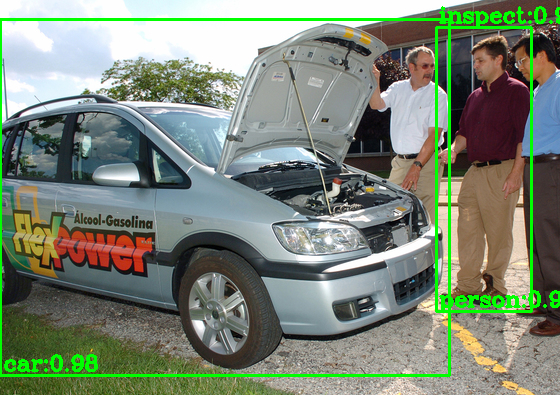} \\
\loadFig{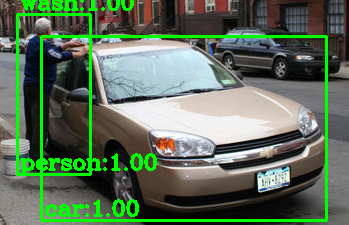} & 
\loadFig{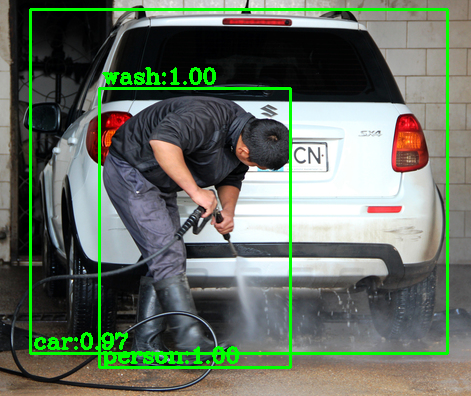} & 
\loadFig{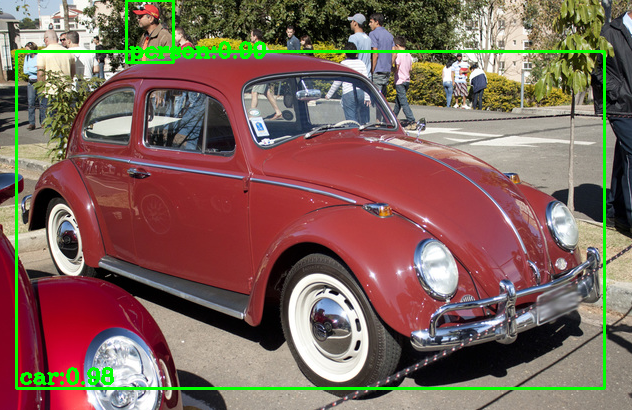} & 
\loadFig{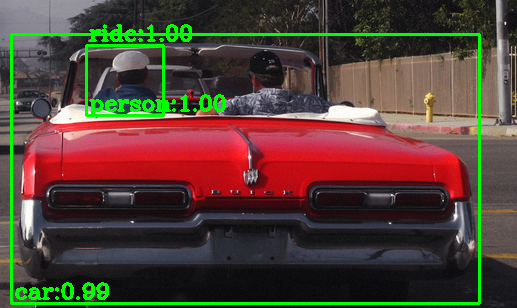} \\
\midrule
\multicolumn{4}{c}{Query images:} \\
\multicolumn{2}{r}{\loadFig{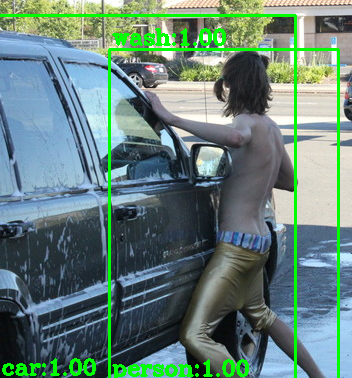}} &
\multicolumn{2}{l}{\loadFig{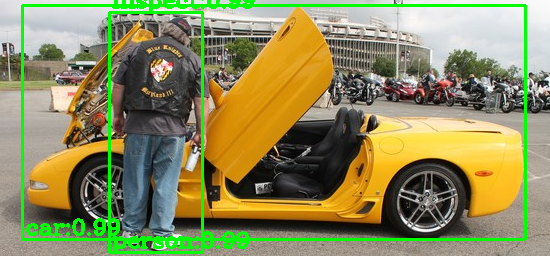}} \\
\multicolumn{4}{c}{Predictions:~~~~~~~~~\textbf{positive}~~~~~~~~~~~~~~~~~~~~~~~~\textbf{negative}}
\end{tabular}
\vskip -0.15in
\caption{\textbf{Illustration of our oracle model.} The concept in $\calP$ is \texttt{wash car}.
}
\label{fig:oracle_1}
\end{figure*}

\begin{figure*}[t]
\centering
\renewcommand{\tabcolsep}{1.5pt}
\newcommand{\loadFig}[1]{\includegraphics[height=0.121\linewidth]{#1}}
\begin{tabular}{cc|cc}
\multicolumn{2}{c|}{$\calP$} & \multicolumn{2}{c}{$\calN$} \\
\loadFig{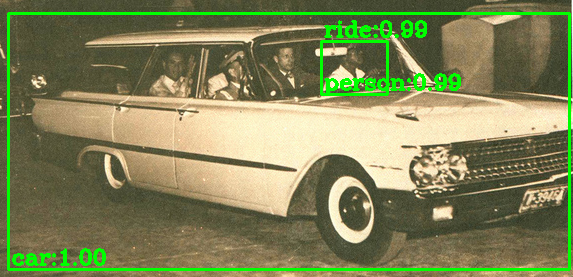} & 
\loadFig{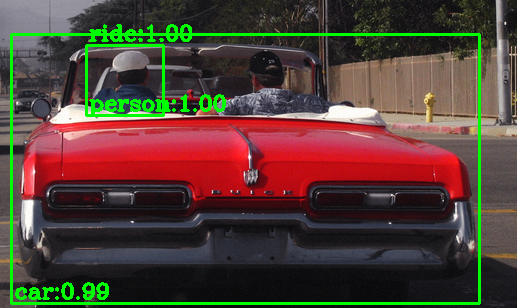} & 
\loadFig{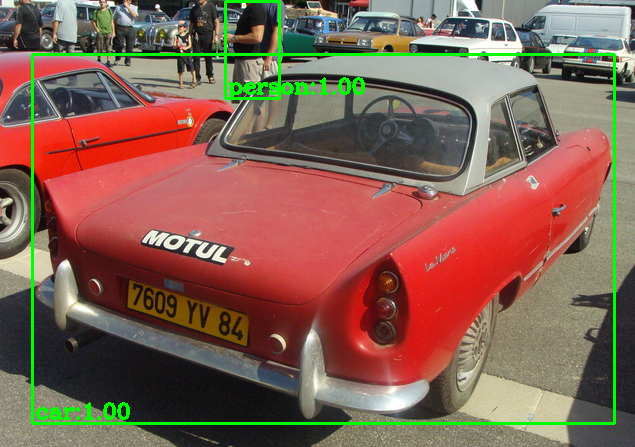} & 
\loadFig{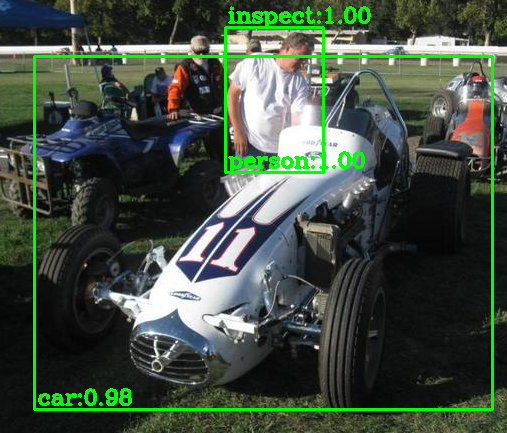} \\
\loadFig{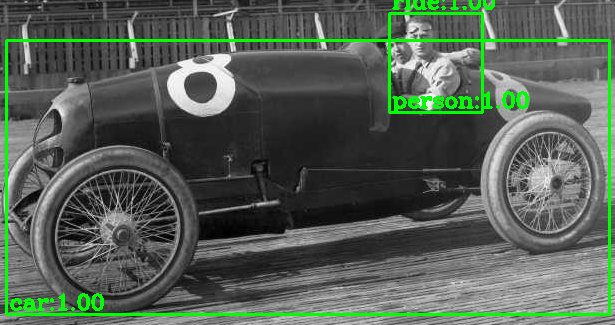} & 
\loadFig{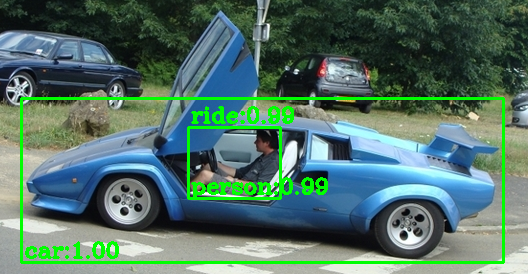} & 
\loadFig{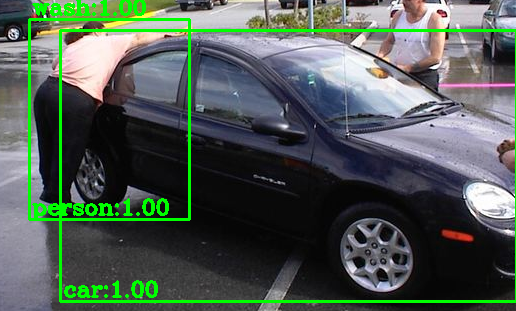} & 
\loadFig{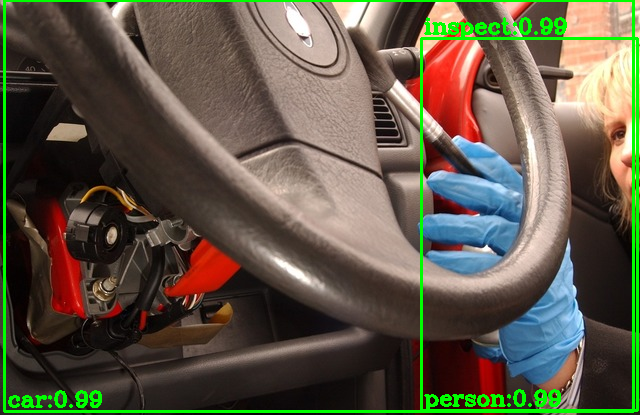} \\
\loadFig{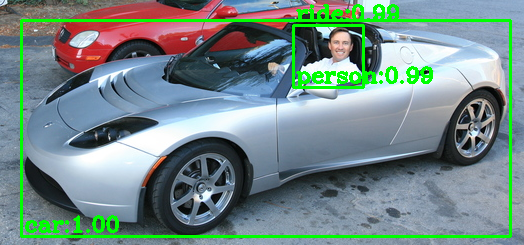} & 
\loadFig{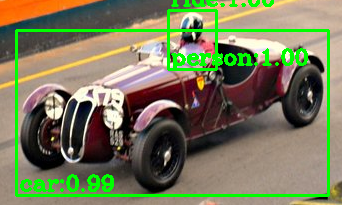} & 
\loadFig{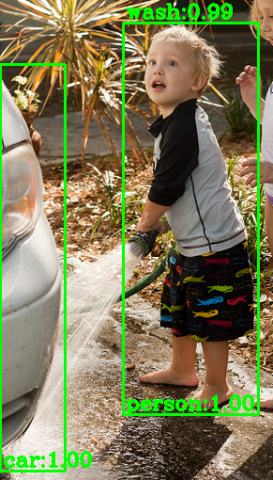} & 
\loadFig{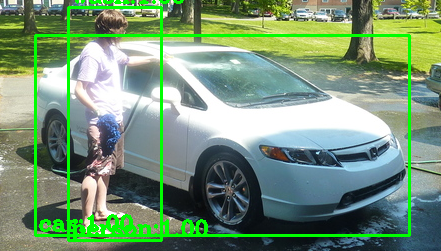} \\
\midrule
\multicolumn{4}{c}{Query images:} \\
\multicolumn{2}{r}{\loadFig{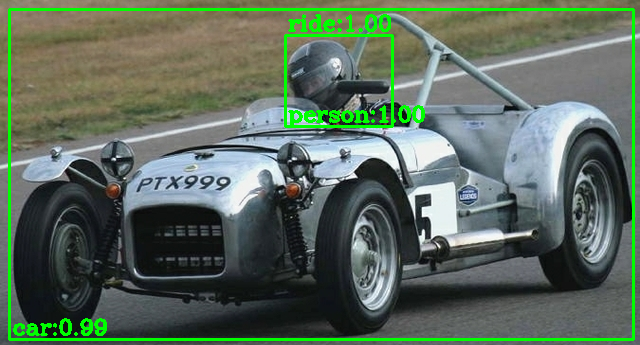}} &
\multicolumn{2}{l}{\loadFig{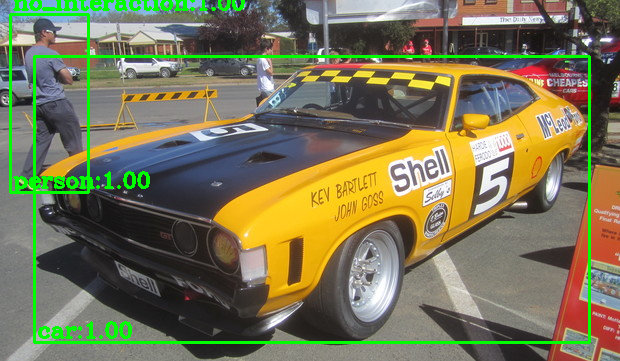}} \\
\multicolumn{4}{c}{Predictions:~~~~~~~~~~~~\textbf{positive}~~~~~~~~~~~~~~~~~~~~~~~~~~~~~~\textbf{negative}~~~~~~~~~~~~~~~~~~~~~}
\end{tabular}
\vskip -0.15in
\caption{\textbf{Illustration of our oracle model.} The concept in $\calP$ is \texttt{ride car}.
}
\label{fig:oracle_2}
\end{figure*}

\begin{figure*}[t]
\centering
\renewcommand{\tabcolsep}{1.5pt}
\newcommand{\loadFig}[1]{\includegraphics[height=0.141\linewidth]{#1}}
\begin{tabular}{cc|cc}
\multicolumn{2}{c|}{$\calP$} & \multicolumn{2}{c}{$\calN$} \\
\loadFig{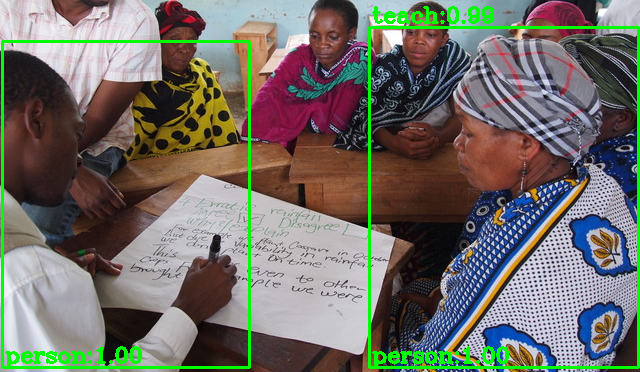} & 
\loadFig{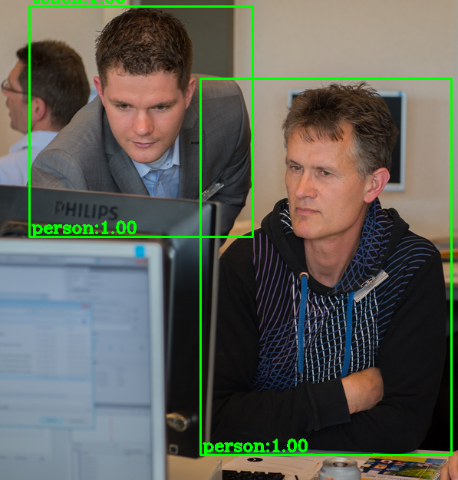} & 
\loadFig{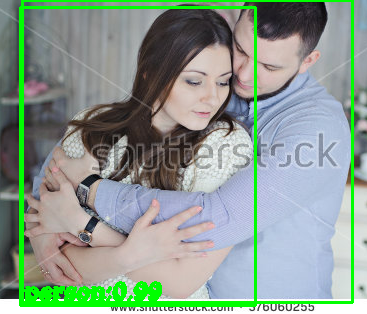} & 
\loadFig{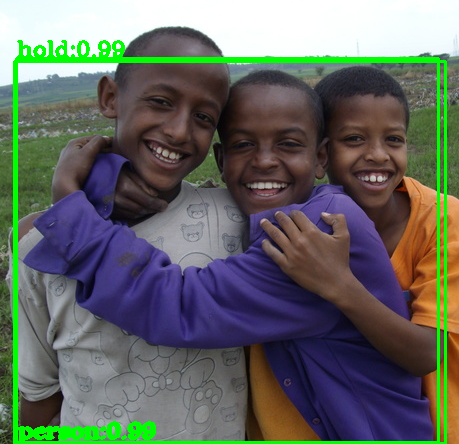} \\
\loadFig{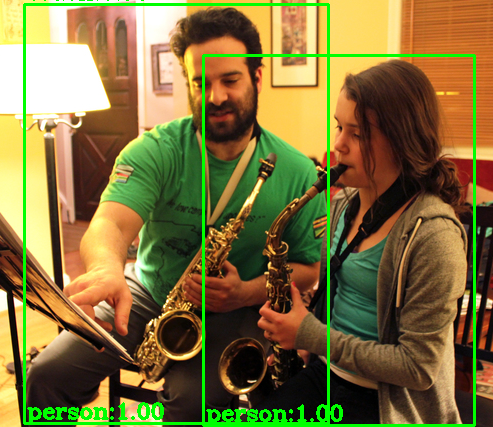} & 
\loadFig{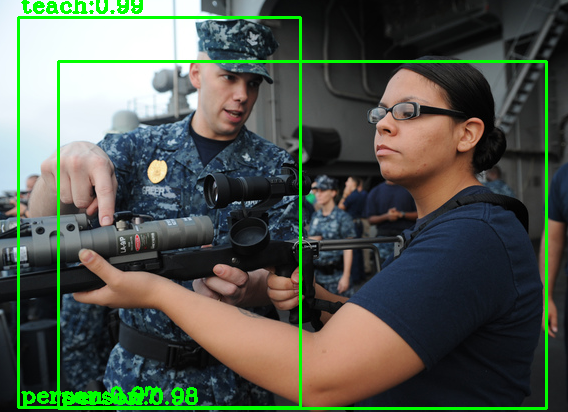} & 
\loadFig{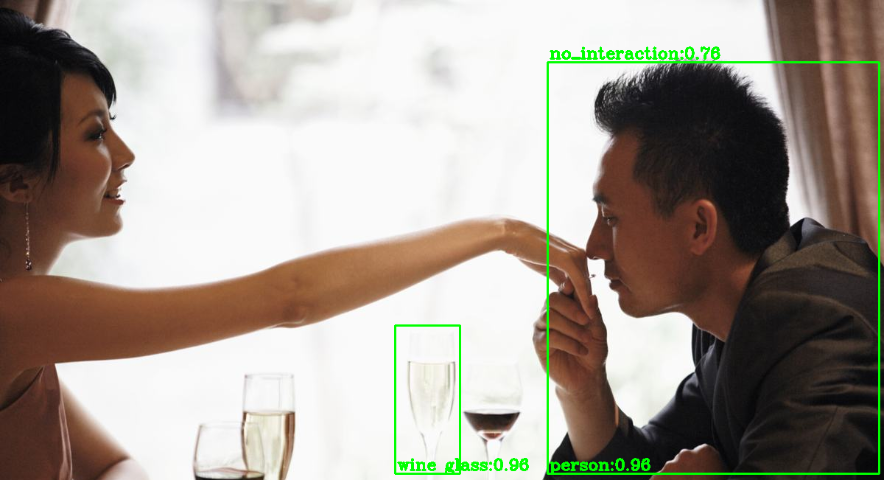} & 
\loadFig{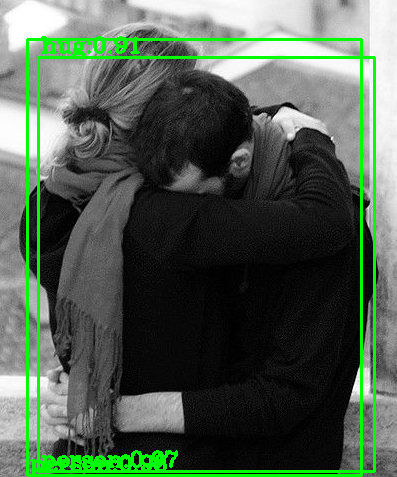} \\
\loadFig{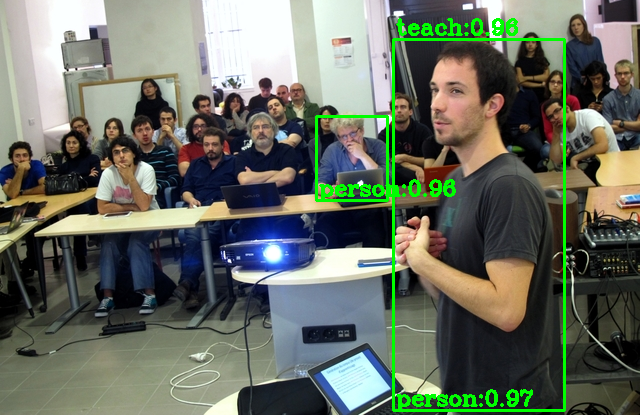} & 
\loadFig{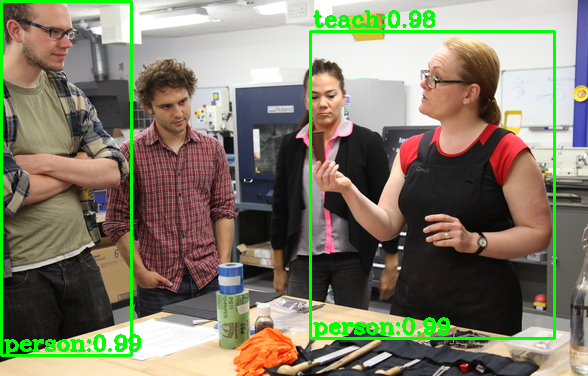} & 
\loadFig{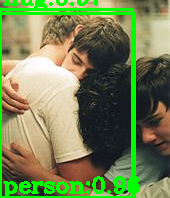} & 
\loadFig{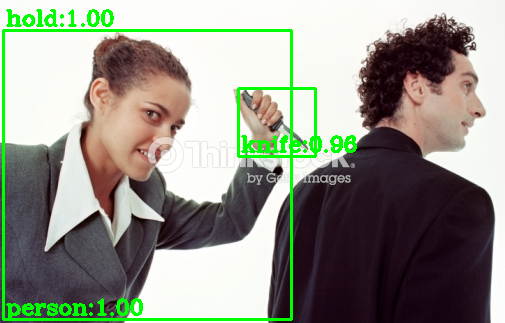} \\
\midrule
\multicolumn{4}{c}{Query images:} \\
\multicolumn{2}{r}{\loadFig{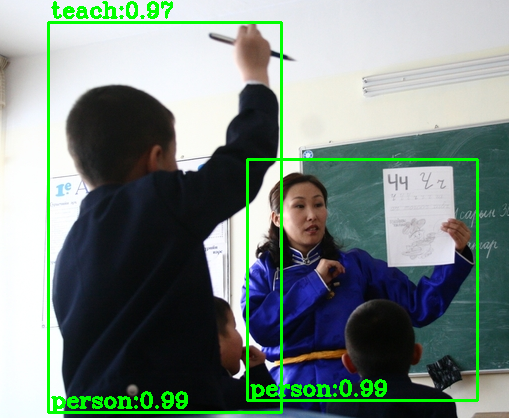}} &
\multicolumn{2}{l}{\loadFig{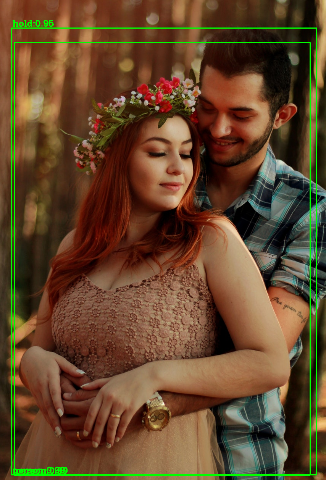}} \\
\multicolumn{4}{c}{Predictions:~~~~~~~~~\textbf{positive}~~~~~~~~~~~\textbf{negative}~~~~~~~~~~~~~~~~~~~~~~~~~~~~~~~~~~~~~}
\end{tabular}
\vskip -0.15in
\caption{\textbf{Illustration of our oracle model.} The concept in $\calP$ is \texttt{teach person}.
}
\label{fig:oracle_2_2}
\end{figure*}

\begin{figure*}[t]
\centering
\renewcommand{\tabcolsep}{1.5pt}
\newcommand{\loadFig}[1]{\includegraphics[height=0.201\linewidth]{#1}}
\begin{tabular}{cc|cc}
\multicolumn{2}{c|}{$\calP$} & \multicolumn{2}{c}{$\calN$} \\
\loadFig{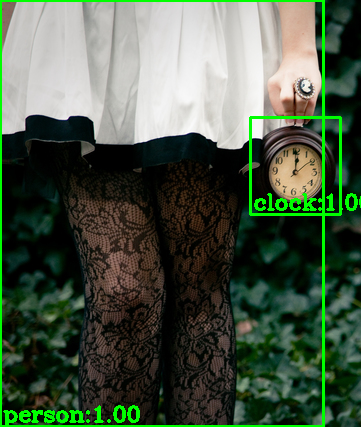} & 
\loadFig{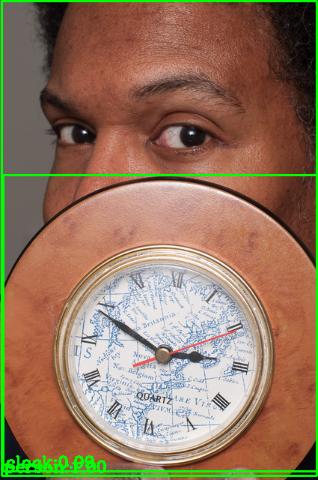} & 
\loadFig{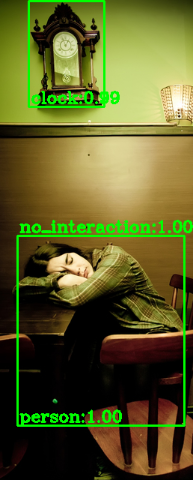} & 
\loadFig{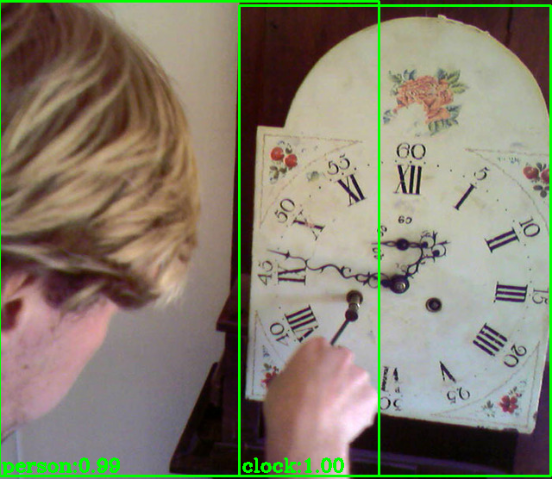} \\
\loadFig{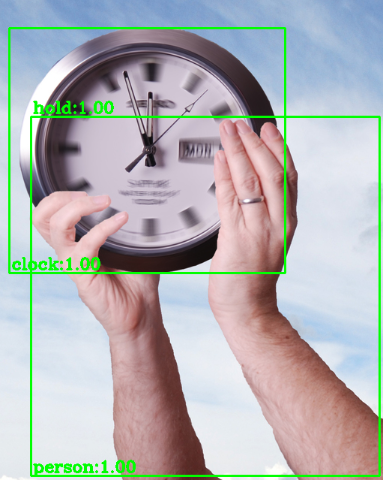} & 
\loadFig{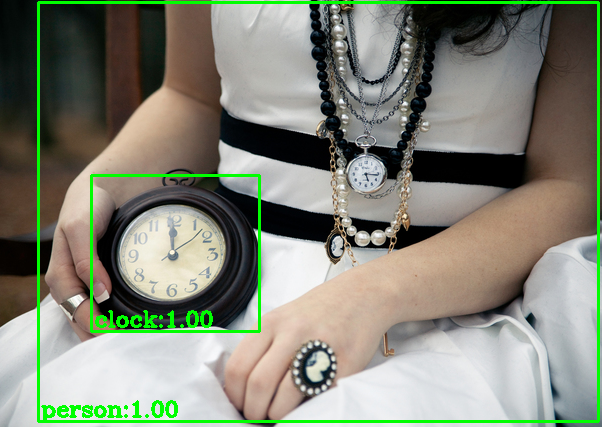} & 
\loadFig{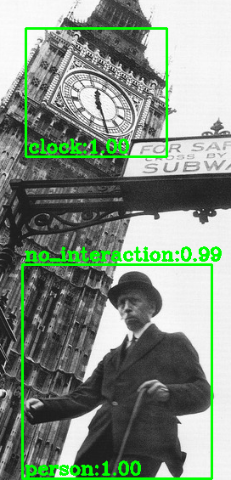} & 
\loadFig{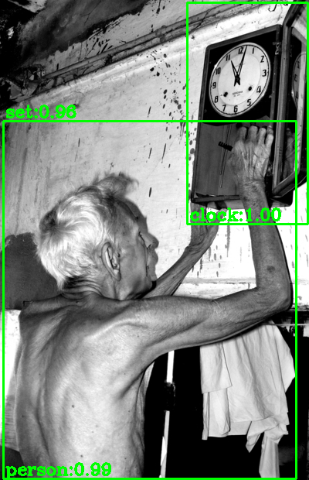} \\
\loadFig{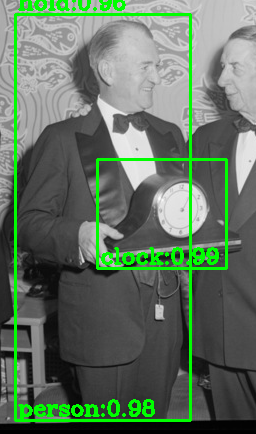} & 
\loadFig{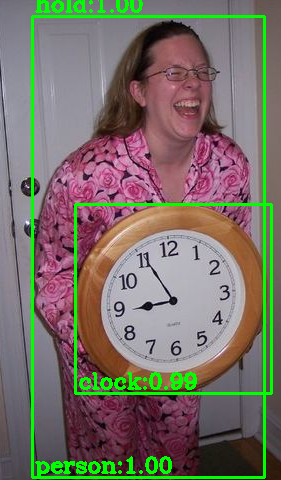} & 
\loadFig{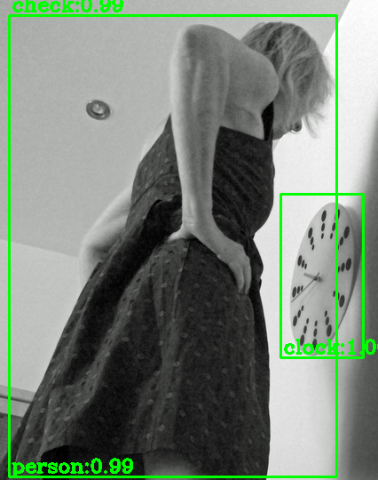} & 
\loadFig{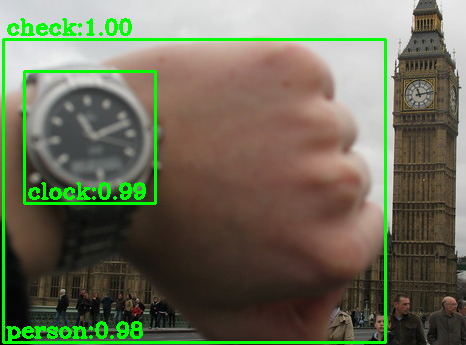} \\
\midrule
\multicolumn{4}{c}{Query images:} \\
\multicolumn{2}{r}{\loadFig{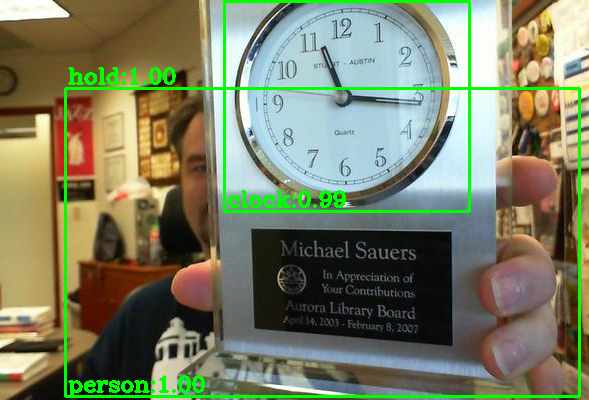}} &
\multicolumn{2}{l}{\loadFig{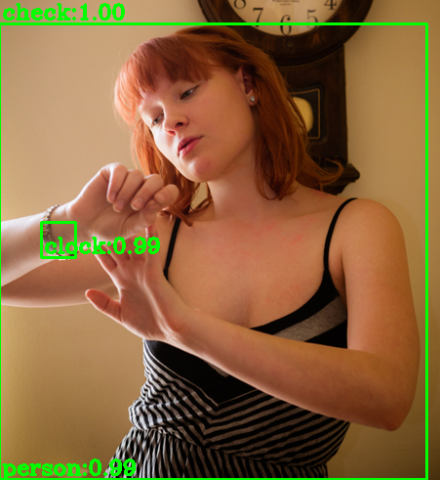}} \\
\multicolumn{4}{c}{Predictions:~~~~~~~~~~~~~~~~~~~~~~~~~~\textbf{positive}~~~~~~~~~~~~~~~~~~~~~~~~~~~~~~~~~\textbf{negative}~~~~~~~~~~~~~~~~~~~~~~~~~~~~~~~~~~~~~}
\end{tabular}
\vskip -0.15in
\caption{\textbf{Illustration of our oracle model.} The concept in $\calP$ is \texttt{hold clock}.
}
\label{fig:oracle_3}
\end{figure*}

\begin{figure*}[t]
\centering
\renewcommand{\tabcolsep}{1.5pt}
\newcommand{\loadFig}[1]{\includegraphics[height=0.201\linewidth]{#1}}
\begin{tabular}{cc|cc}
\multicolumn{2}{c|}{$\calP$} & \multicolumn{2}{c}{$\calN$} \\
\loadFig{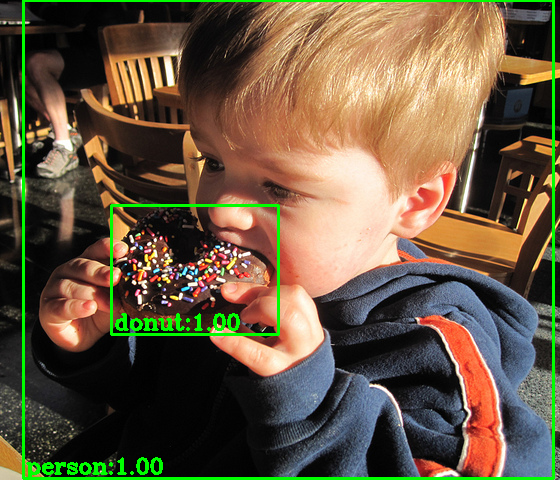} & 
\loadFig{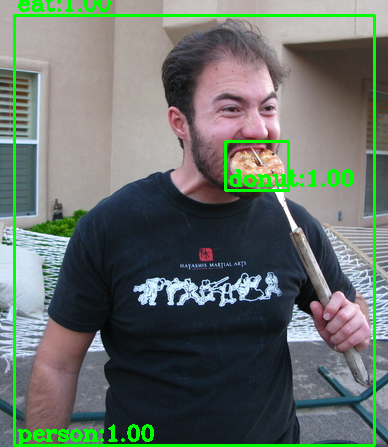} & 
\loadFig{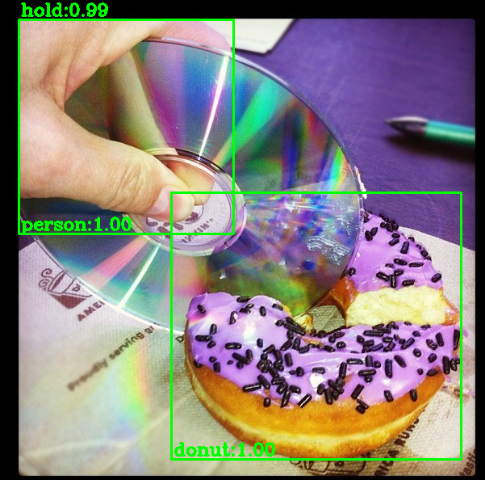} & 
\loadFig{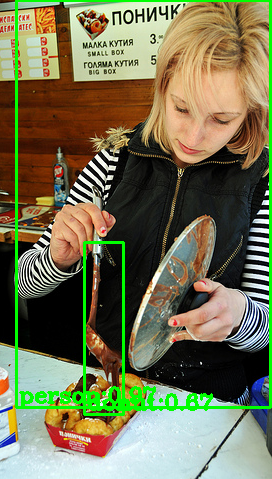} \\
\loadFig{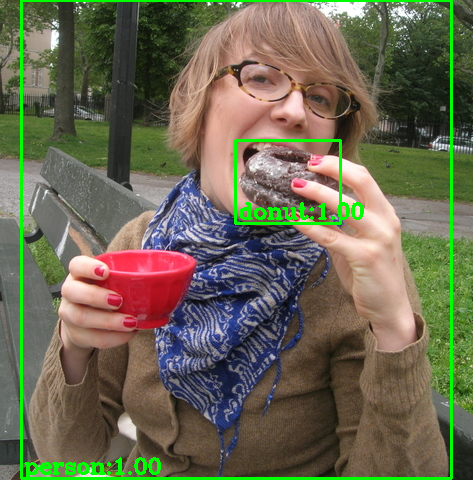} & 
\loadFig{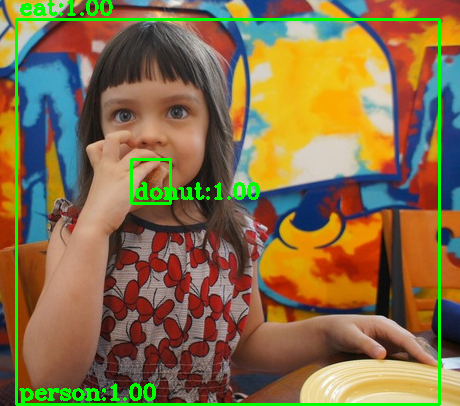} & 
\loadFig{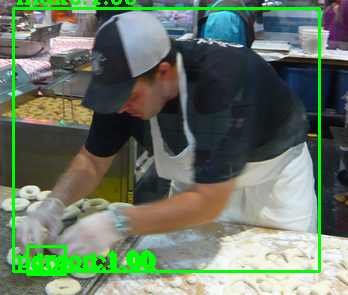} & 
\loadFig{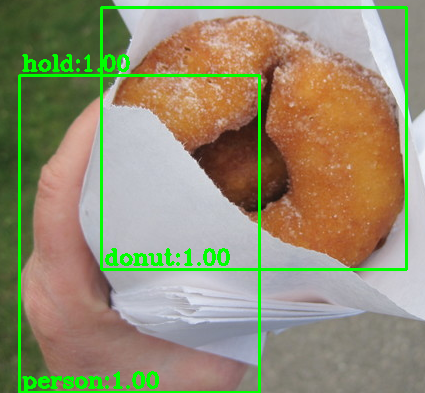} \\
\loadFig{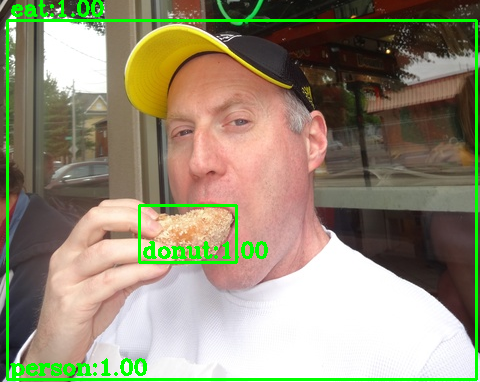} & 
\loadFig{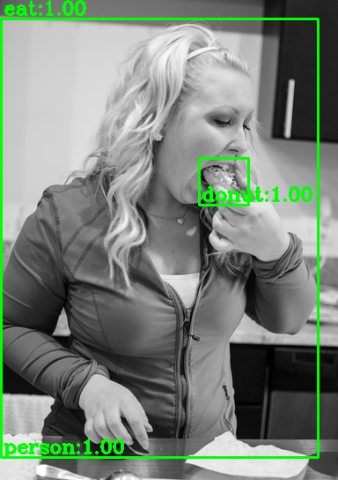} & 
\loadFig{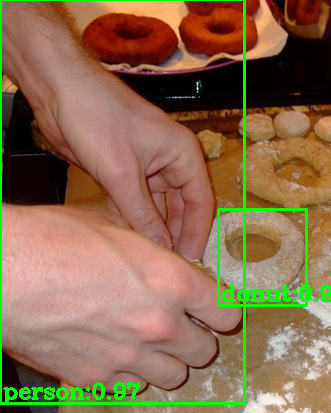} & 
\loadFig{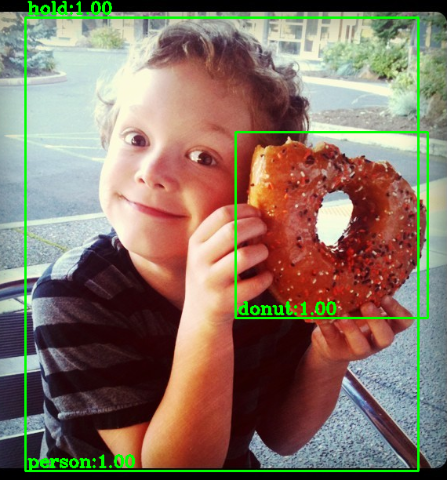} \\
\midrule
\multicolumn{4}{c}{Query images:} \\
\multicolumn{2}{r}{\loadFig{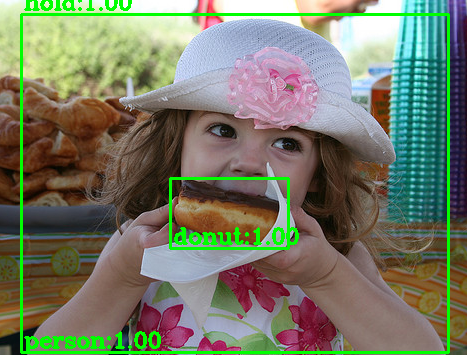}} &
\multicolumn{2}{l}{\loadFig{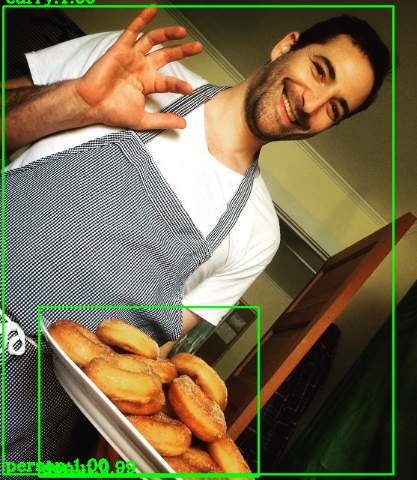}} \\
\multicolumn{4}{c}{~Predictions:~~~~~~~~~~~~~~~~~~~\textbf{negative (\textcolor{red}{wrong})}~~~~~~~~~~~~~~~~~~~~~~~\textbf{negative}~~~~~~~~~~~~~~~~~~~~~~~~~~~~~~~~~~~~~~~~~~~~~~~~~~~~~~~~~~~~~~~~~~~~~~}
\end{tabular}
\vskip -0.15in
\caption{\textbf{A failure of our oracle model.} The concept in $\calP$ is \texttt{eat cake}. The \emph{HOITrans} model~\cite{zou2021end} incorrectly recognizes the first query image as \texttt{hold cake} (which should be \texttt{eat cake}). As a result, it makes a wrong prediction for the first query image.
}
\label{fig:oracle_5}
\end{figure*}

\section{More details on the oracle model}
We first review how our oracle model works. Denoting the HOI detections in the $\calP$ and $\calN$ as $\calD^P$ and $\calD^N$, respectively. $\calD^P$ contains the detections from all of the images in the $\calP$, defined as $\calD^P=\{c_i^P\}_{i=1}^{N_P}$, where $c_i^P$ is a HOI triplet. $N_P$ is the total number of detections. Note that there may be multiple or no detections for a single image. Similarly, $\calD^P$ is defined as $\calD^P=\{c_i^P\}_{i=1}^{N_P}$. According to the property of Bongard-HOI, the visual concept $c_P$ should only appear in the $\calP$, not in the $\calN$. We, therefore, compute $c_P$ as
\begin{align*}
    c_P = \texttt{majority\_vote}(\calD^P - \calD^N),
\end{align*}
where $-$ is the set operator for set subtraction. Given the detections $\calD^q=\{c_i^q\}_{i=1}^{N_q}$ for the query image $I_q$, our prediction $y$ becomes
\begin{align*}
  y=\left\{\begin{array}{cc}
      1, & \textrm{if}~c_P \in \calD^q, \\
      0, & \textrm{otherwise.}
  \end{array} \right.
\end{align*}

We now discuss some possible corner cases where the main paper does not cover.
\vskip 0.05in
\noindent\textbf{What if} \texttt{majority\_vote} \textbf{return multiple concepts?} In this case, we simply enumerate each of them when making predictions for $y$. The predicted $y$ will be $1$ as long as at least one returned concepts present in $\calD^q$; otherwise it will be $0$.

\vskip 0.05in
\noindent\textbf{What if} $\calD^P$, $\calD^N$ or $\calD^q$ \textbf{is empty?} In case when $\calD^P$ is empty, we view this example as an failure case for our oracle model, as it does not induce the right concept as expected. On the contrary, it's totally fine that $\calD^N$, meaning that no detection need to be removed from $\calD^P$. Finally, how we handle the case when $\calD_q$ is empty depends on the true label $y^\star$. If $y^\star$ is $1$, then we view this example as an failure case. But we will make the prediction an automatic success if $y^\star$ is 0, since our oracle model finds there is no ground truth concept presenting in the query, which should be the right prediction.

We show successful cases of our oracle model in Fig.~\ref{fig:oracle_1}, Fig.~\ref{fig:oracle_2}, Fig.~\ref{fig:oracle_2_2}, Fig.~\ref{fig:oracle_3}. A failure case is shown in Fig.~\ref{fig:oracle_5}. 

\section{More Details on MTurk Data Curation}
\label{sec:mturk}
\begin{figure*}
\newcommand{\loadFig}[1]{\includegraphics[width=0.7\linewidth]{#1}}
    \centering
    \loadFig{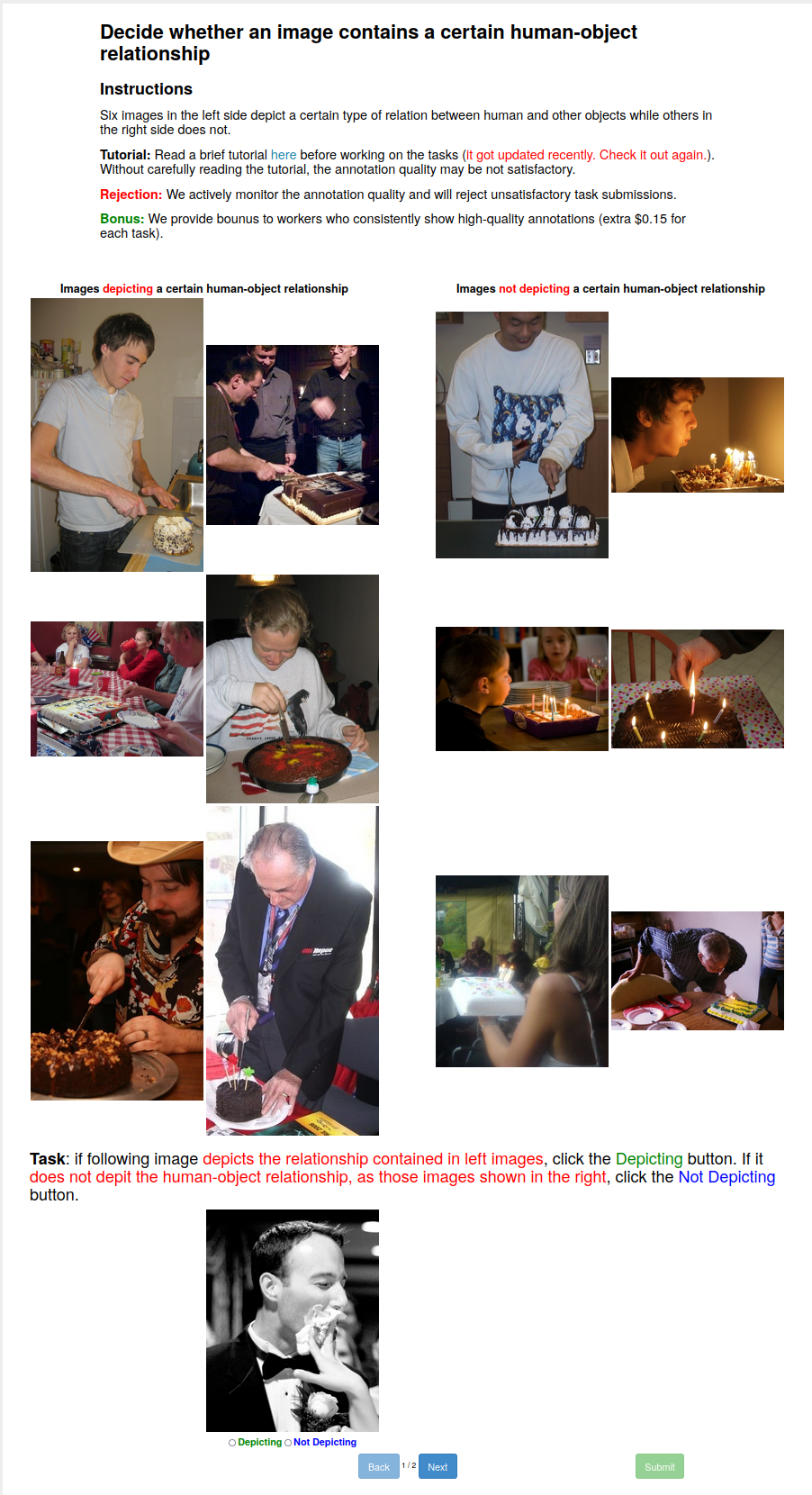}
    \caption{\textbf{The user interface (UI) of MTurk data curation.}}
    \label{fig:mturk_ui}
\end{figure*}

\noindent\textbf{User interface.} The user interface of data curation on the Amazon Mechanical Turkp (MTurk) platform is shown in Fig.~\ref{fig:mturk_ui}. In the top part, we show images depicting a common visual relationship between human and objects in the left (\ie, positive examples $\calP$ in our Bongard problem). In the right, images that do not contain the visual relationship are shown (\ie, negative examples $\calN$). In the bottom part, given a query image, a tester needs to decide whether it depicts the particular visual relationship or not. Each MTurk job contains two few-shot instances, where a tester can freely switch between two pages. They can only submit the job once both two tasks are finished.

We do not tell the testers what objects to focus on to induce the common visual relationship. It is intended to be similar to what a machine learning model does, which needs to do object detection first. 

\begin{figure*}
\newcommand{\loadFig}[1]{\includegraphics[width=0.52\linewidth]{#1}}
\centering
\loadFig{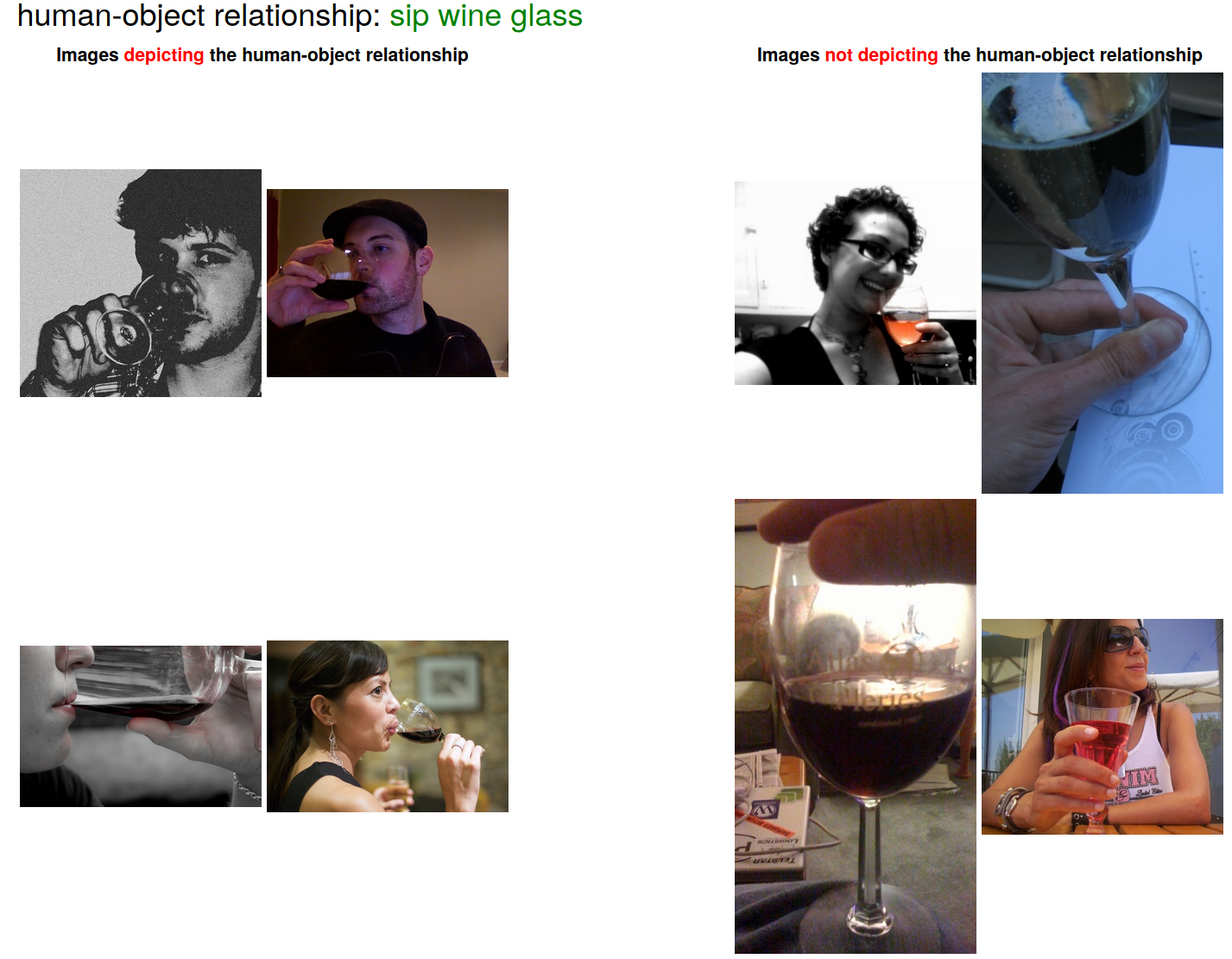}\\
\loadFig{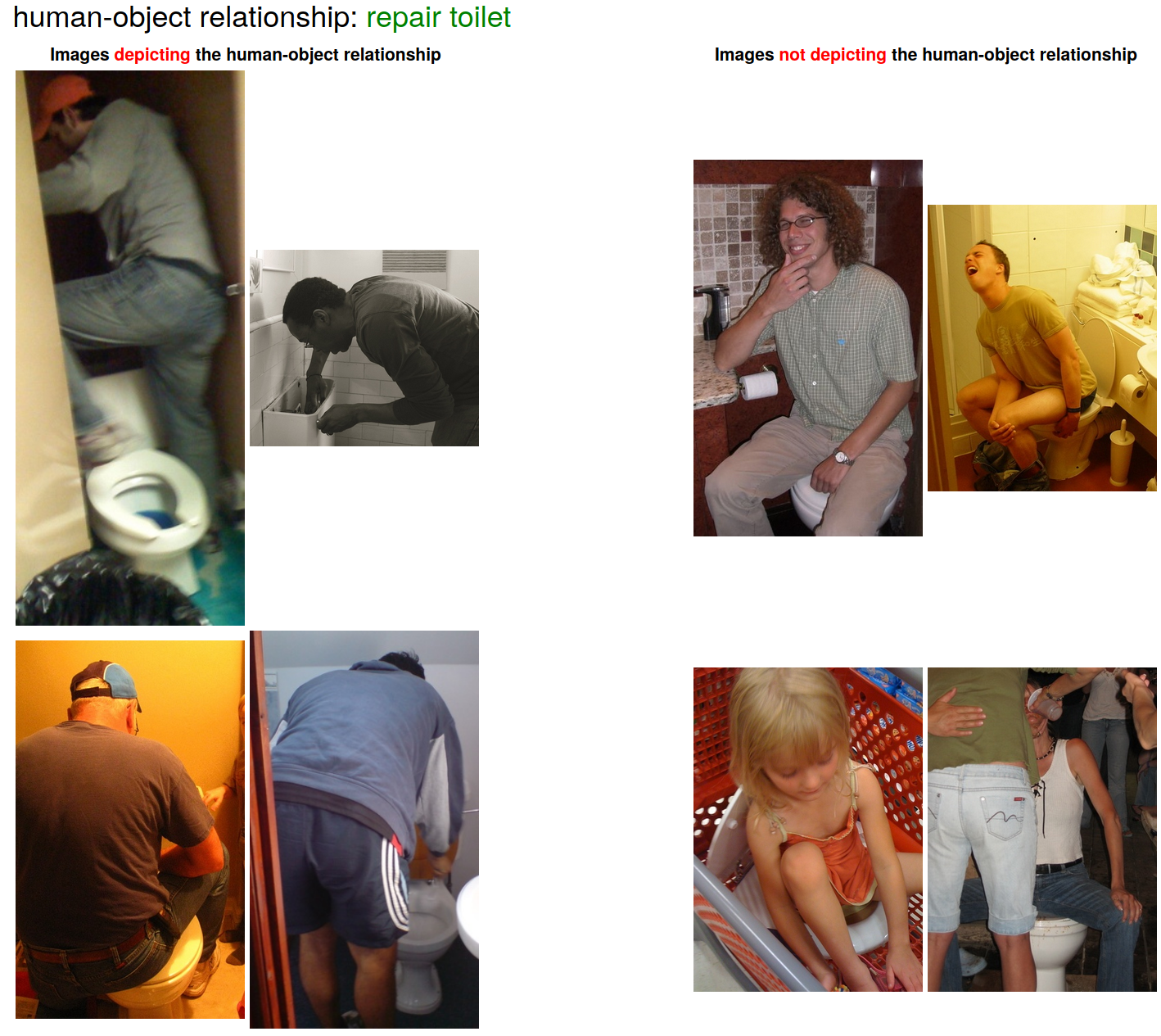}\\
\loadFig{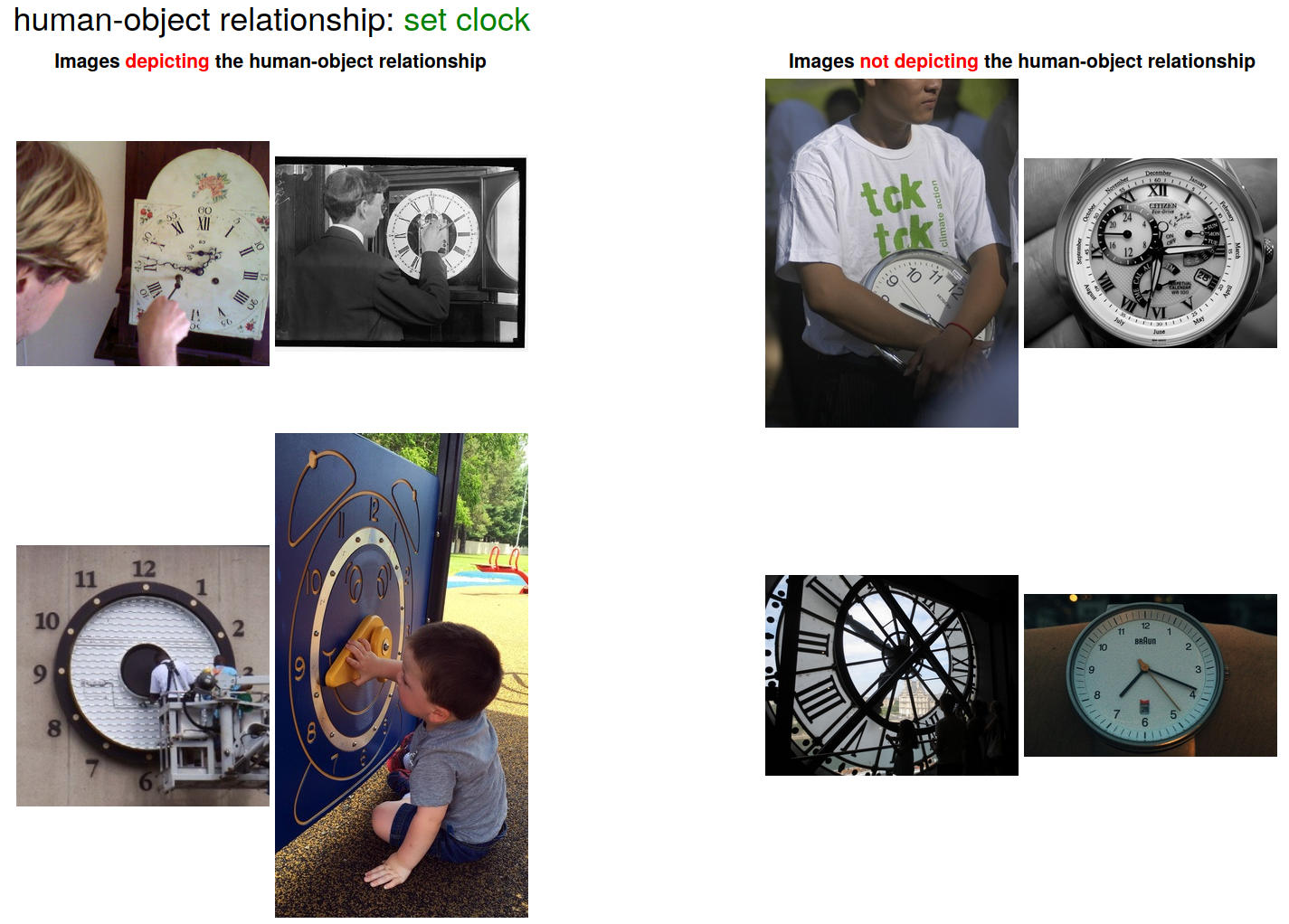}\\
\caption{\textbf{Examples of different visual relationships given to MTurk testers.} For each example, we tell what the visual relationship is so that the testers can better understand the scope of the job.}
\label{fig:mturk_ex}
\end{figure*}

\noindent\textbf{Simple examples given to testers.} To ensure testers who see the form of few-shot instances for the first time can successfully finish the job, we provide some examples of different visual relationships and encourage them to take a look at these examples before starting working on a job. Such examples are shown in Fig.~\ref{fig:mturk_ex}.

\noindent\textbf{MTurk job setting.} We provided more details about the job setting below.
\begin{itemize}
\item \textbf{Region.} We restrict the regions of testers to be in the US and Germany.
\item \textbf{Approval rate.} Each MTurker tester maintains a job approval rate based on their performance on previous jobs. We invite only MTurk testers whose job approval rate is equal to or greater than 98\%.
\item \textbf{Number of approved jobs.} Setting a qualification for the job approval rate only is not sufficient to hire high-quality testers since newly registered novel testers have a job approval rate of 100\%. Therefore, we also set a qualification such that only testers who have more than 500 jobs approved previously are invited.
\item \textbf{Invited annotators.} Through a couple of small-scale preliminary studies, we identified 35 reliable annotators on MTurk. For the large-scale data curation, we invited them to participate only.
\item \textbf{Reward setting.} We provide \$0.15 for each job with an additional \$0.15 bonus if consistently high-quality annotations are made. According to our experiences of finishing the job, it roughly corresponds to about \$30 per hour.
\item \textbf{Number of testers for each job.} We hire three independent testers for each job and aggregate their annotations. In specific, we only keep the few-shot instances where at least one of the three testers correctly classified the query image according to the ground-truth annotations. Otherwise, it suggests that a BP is either ambiguous or too difficult. We discard 2.5\% of the few-shot instances that we submitted to MTurk.
\item \textbf{Job life time.} A job will not be available after 7 days if it is not claimed by any tester. But we found that all of the jobs were finished within such a limit.
\end{itemize}

{
\small 
\bibliographystyle{ieee_fullname}
\bibliography{egbib}

\begin{thebibliography}{10}\itemsep=-1pt

\bibitem{bakhtin2019phyre}
Anton Bakhtin, Laurens van~der Maaten, Justin Johnson, Laura Gustafson, and
  Ross Girshick.
\newblock Phyre: A new benchmark for physical reasoning.
\newblock In {\em Advances in Neural Information Processing Systems}, pages
  5083--5094, 2019.

\bibitem{barrett2018measuring}
David Barrett, Felix Hill, Adam Santoro, Ari Morcos, and Timothy Lillicrap.
\newblock Measuring abstract reasoning in neural networks.
\newblock In {\em ICML}, pages 511--520, 2018.

\bibitem{bongard1968recognition}
Mikhail~Moiseevich Bongard.
\newblock The recognition problem.
\newblock Technical report, Foreign Technology Div Wright-Patterson AFB Ohio,
  1968.

\bibitem{chao2015hico}
Yu{-}Wei Chao, Zhan Wang, Yugeng He, Jiaxuan Wang, and Jia Deng.
\newblock {HICO:} {A} benchmark for recognizing human-object interactions in
  images.
\newblock In {\em ICCV}, 2015.

\bibitem{chen2020improved}
Xinlei Chen, Haoqi Fan, Ross~B. Girshick, and Kaiming He.
\newblock Improved baselines with momentum contrastive learning.
\newblock {\em CoRR}, abs/2003.04297, 2020.

\bibitem{chen2020new}
Yinbo Chen, Xiaolong Wang, Zhuang Liu, Huijuan Xu, and Trevor Darrell.
\newblock A new meta-baseline for few-shot learning.
\newblock {\em arXiv preprint arXiv:2003.04390}, 2020.

\bibitem{chollet2019measure}
Fran{\c{c}}ois Chollet.
\newblock On the measure of intelligence.
\newblock {\em arXiv preprint arXiv:1911.01547}, 2019.

\bibitem{deng2009imagenet}
Jia Deng, Wei Dong, Richard Socher, Li-Jia Li, Kai Li, and Li Fei-Fei.
\newblock Imagenet: A large-scale hierarchical image database.
\newblock In {\em 2009 IEEE conference on computer vision and pattern
  recognition}, pages 248--255. IEEE, 2009.

\bibitem{fe2003bayesian}
Li Fe-Fei et~al.
\newblock A bayesian approach to unsupervised one-shot learning of object
  categories.
\newblock In {\em Proceedings Ninth IEEE International Conference on Computer
  Vision}, pages 1134--1141. IEEE, 2003.

\bibitem{gupta2015visual}
Saurabh Gupta and Jitendra Malik.
\newblock Visual semantic role labeling.
\newblock {\em arXiv preprint arXiv:1505.04474}, 2015.

\bibitem{he2020mask}
Kaiming He, Georgia Gkioxari, Piotr Doll{\'{a}}r, and Ross~B. Girshick.
\newblock Mask {R-CNN}.
\newblock {\em {IEEE} TPAMI}, 42(2):386--397, 2020.

\bibitem{he2016deep}
Kaiming He, Xiangyu Zhang, Shaoqing Ren, and Jian Sun.
\newblock Deep residual learning for image recognition.
\newblock In {\em CVPR}, pages 770--778, 2016.

\bibitem{hochreiter2001learning}
Sepp Hochreiter, A~Steven Younger, and Peter~R Conwell.
\newblock Learning to learn using gradient descent.
\newblock In {\em International Conference on Artificial Neural Networks},
  pages 87--94. Springer, 2001.

\bibitem{hou2021fcl}
Zhi Hou, Yu Baosheng, Yu Qiao, Xiaojiang Peng, and Dacheng Tao.
\newblock Detecting human-object interaction via fabricated compositional
  learning.
\newblock In {\em CVPR}, 2021.

\bibitem{hou2020visual}
Zhi Hou, Xiaojiang Peng, Yu Qiao, and Dacheng Tao.
\newblock Visual compositional learning for human-object interaction detection.
\newblock In {\em Proceedings of the European Conference on Computer Vision
  (ECCV)}, pages 584--600. Springer, 2020.

\bibitem{johnson2017clevr}
Justin Johnson, Bharath Hariharan, Laurens van~der Maaten, Li Fei{-}Fei,
  C.~Lawrence Zitnick, and Ross~B. Girshick.
\newblock {CLEVR:} {A} diagnostic dataset for compositional language and
  elementary visual reasoning.
\newblock In {\em CVPR}, 2017.

\bibitem{kato2018compositional}
Keizo Kato, Yin Li, and Abhinav Gupta.
\newblock Compositional learning for human object interaction.
\newblock In {\em Proceedings of the European Conference on Computer Vision
  (ECCV)}, pages 234--251. Springer, 2018.

\bibitem{koch2015siamese}
Gregory Koch, Richard Zemel, and Ruslan Salakhutdinov.
\newblock Siamese neural networks for one-shot image recognition.
\newblock In {\em ICML deep learning workshop}, volume~2. Lille, 2015.

\bibitem{krishna2017visual}
Ranjay Krishna, Yuke Zhu, Oliver Groth, Justin Johnson, Kenji Hata, Joshua
  Kravitz, Stephanie Chen, Yannis Kalantidis, Li{-}Jia Li, David~A. Shamma,
  Michael~S. Bernstein, and Li Fei{-}Fei.
\newblock Visual genome: Connecting language and vision using crowdsourced
  dense image annotations.
\newblock {\em IJCV}, 123(1):32--73, 2017.

\bibitem{OpenImages}
Alina Kuznetsova, Hassan Rom, Neil Alldrin, Jasper Uijlings, Ivan Krasin, Jordi
  Pont-Tuset, Shahab Kamali, Stefan Popov, Matteo Malloci, Alexander
  Kolesnikov, Tom Duerig, and Vittorio Ferrari.
\newblock The open images dataset v4: Unified image classification, object
  detection, and visual relationship detection at scale.
\newblock {\em IJCV}, 2020.

\bibitem{lake2011one}
Brenden~M. Lake, Ruslan Salakhutdinov, Jason Gross, and Joshua~B. Tenenbaum.
\newblock One shot learning of simple visual concepts.
\newblock In Laura~A. Carlson, Christoph H{\"{o}}lscher, and Thomas~F. Shipley,
  editors, {\em CogSci}, 2011.

\bibitem{lake2015human}
Brenden~M Lake, Ruslan Salakhutdinov, and Joshua~B Tenenbaum.
\newblock Human-level concept learning through probabilistic program induction.
\newblock {\em Science}, 350(6266):1332--1338, 2015.

\bibitem{lee2019meta}
Kwonjoon Lee, Subhransu Maji, Avinash Ravichandran, and Stefano Soatto.
\newblock Meta-learning with differentiable convex optimization.
\newblock In {\em Proceedings of the IEEE Conference on Computer Vision and
  Pattern Recognition}, pages 10657--10665, 2019.

\bibitem{li2019hake}
Yonglu Li, Liang Xu, Xijie Huang, Xinpeng Liu, Ze Ma, Mingyang Chen, Shiyi
  Wang, Haoshu Fang, and Cewu Lu.
\newblock {HAKE:} human activity knowledge engine.
\newblock {\em CoRR}, abs/1904.06539, 2019.

\bibitem{li2020detailed}
Yong-Lu Li, Xinpeng Liu, Han Lu, Shiyi Wang, Junqi Liu, Jiefeng Li, and Cewu
  Lu.
\newblock Detailed 2d-3d joint representation for human-object interaction.
\newblock In {\em CVPR}, 2020.

\bibitem{liao2020ppdm}
Yue Liao, Si Liu, Fei Wang, Yanjie Chen, Chen Qian, and Jiashi Feng.
\newblock {PPDM:} parallel point detection and matching for real-time
  human-object interaction detection.
\newblock In {\em CVPR}, 2020.

\bibitem{lin2014microsoft}
Tsung-Yi Lin, Michael Maire, Serge Belongie, James Hays, Pietro Perona, Deva
  Ramanan, Piotr Doll{\'a}r, and C~Lawrence Zitnick.
\newblock Microsoft coco: Common objects in context.
\newblock In {\em European conference on computer vision}, pages 740--755.
  Springer, 2014.

\bibitem{liu2015faceattributes}
Ziwei Liu, Ping Luo, Xiaogang Wang, and Xiaoou Tang.
\newblock Deep learning face attributes in the wild.
\newblock In {\em ICCV}, December 2015.

\bibitem{lu2016visual}
Cewu Lu, Ranjay Krishna, Michael~S. Bernstein, and Fei{-}Fei Li.
\newblock Visual relationship detection with language priors.
\newblock In {\em ECCV}, 2016.

\bibitem{ma2022relvit}
Xiaojian Ma, Weili Nie, Zhiding Yu, Huaizu Jiang, Chaowei Xiao, Yuke Zhu,
  Song-Chun Zhu, and Anima Anandkumar.
\newblock Relvit: Concept-guided vision transformer for visual relational
  reasoning.
\newblock In {\em International Conference on Learning Representations}, 2022.

\bibitem{massiceti2021orbit}
Daniela Massiceti, Luisa Zintgraf, John Bronskill, Lida Theodorou, Matthew
  Tobias~Harris, Ed Cutrell, Cecily Morrison, Katja Hofmann, and Simone Stumpf.
\newblock Orbit: A real-world few-shot dataset for teachable object
  recognition.
\newblock In {\em ICCV}, 2021.

\bibitem{mishra2018simple}
Nikhil Mishra, Mostafa Rohaninejad, Xi Chen, and Pieter Abbeel.
\newblock A simple neural attentive meta-learner.
\newblock {\em ICLR}, 2018.

\bibitem{nie2020bongard}
Weili Nie, Zhiding Yu, Lei Mao, Ankit~B. Patel, Yuke Zhu, and Anima Anandkumar.
\newblock Bongard-logo: {A} new benchmark for human-level concept learning and
  reasoning.
\newblock In {\em NeurIPS}, 2020.

\bibitem{patterson2016coco}
Genevieve Patterson and James Hays.
\newblock Coco attributes: Attributes for people, animals, and objects.
\newblock {\em European Conference on Computer Vision}, 2016.

\bibitem{raghu2019rapid}
Aniruddh Raghu, Maithra Raghu, Samy Bengio, and Oriol Vinyals.
\newblock Rapid learning or feature reuse? towards understanding the
  effectiveness of maml.
\newblock {\em ICLR}, 2020.

\bibitem{ragusa2021meccano}
Francesco Ragusa, Antonino Furnari, Salvatore Livatino, and Giovanni~Maria
  Farinella.
\newblock The meccano dataset: Understanding human-object interactions from
  egocentric videos in an industrial-like domain.
\newblock In {\em WACV}, pages 1569--1578, 2021.

\bibitem{ren2018meta}
Mengye Ren, Eleni Triantafillou, Sachin Ravi, Jake Snell, Kevin Swersky,
  Joshua~B Tenenbaum, Hugo Larochelle, and Richard~S Zemel.
\newblock Meta-learning for semi-supervised few-shot classification.
\newblock {\em ICLR}, 2018.

\bibitem{ren2017faster}
Shaoqing Ren, Kaiming He, Ross~B. Girshick, and Jian Sun.
\newblock Faster {R-CNN:} towards real-time object detection with region
  proposal networks.
\newblock {\em {IEEE} TPAMI}, 39(6):1137--1149, 2017.

\bibitem{ronchi2015describing}
Matteo~Ruggero Ronchi and Pietro Perona.
\newblock Describing common human visual actions in images.
\newblock In {\em BMVC}, 2015.

\bibitem{santoro2016meta}
Adam Santoro, Sergey Bartunov, Matthew Botvinick, Daan Wierstra, and Timothy
  Lillicrap.
\newblock Meta-learning with memory-augmented neural networks.
\newblock In {\em International conference on machine learning}, pages
  1842--1850, 2016.

\bibitem{santoro2017a}
Adam Santoro, David Raposo, David~G Barrett, Mateusz Malinowski, Razvan
  Pascanu, Peter Battaglia, and Timothy Lillicrap.
\newblock A simple neural network module for relational reasoning.
\newblock In {\em {NeurIPS}}, 2017.

\bibitem{saxton2019analysing}
David Saxton, Edward Grefenstette, Felix Hill, and Pushmeet Kohli.
\newblock Analysing mathematical reasoning abilities of neural models.
\newblock {\em arXiv preprint arXiv:1904.01557}, 2019.

\bibitem{snell17protonet}
Jake Snell, Kevin Swersky, and Richard~S. Zemel.
\newblock {Prototypical Networks for Few-shot Learning}.
\newblock {\em Advances in Neural Information Processing Systems}, 2017.

\bibitem{teney2019v}
Damien Teney, Peng Wang, Jiewei Cao, Lingqiao Liu, Chunhua Shen, and Anton
  van~den Hengel.
\newblock V-prom: A benchmark for visual reasoning using visual progressive
  matrices.
\newblock In {\em AAAI}, 2020.

\bibitem{triantafillou2019meta}
Eleni Triantafillou, Tyler Zhu, Vincent Dumoulin, Pascal Lamblin, Utku Evci,
  Kelvin Xu, Ross Goroshin, Carles Gelada, Kevin Swersky, Pierre-Antoine
  Manzagol, et~al.
\newblock Meta-dataset: A dataset of datasets for learning to learn from few
  examples.
\newblock In {\em ICLR}, 2020.

\bibitem{vaswani2017attention}
Ashish Vaswani, Noam Shazeer, Niki Parmar, Jakob Uszkoreit, Llion Jones,
  Aidan~N. Gomez, Lukasz Kaiser, and Illia Polosukhin.
\newblock Attention is all you need.
\newblock In {\em NeurIPS}, 2017.

\bibitem{vinyals2016matching}
Oriol Vinyals, Charles Blundell, Timothy Lillicrap, Daan Wierstra, et~al.
\newblock Matching networks for one shot learning.
\newblock In {\em NeurIPS}, pages 3630--3638, 2016.

\bibitem{WahCUB_200_2011}
C. Wah, S. Branson, P. Welinder, P. Perona, and S. Belongie.
\newblock {The Caltech-UCSD Birds-200-2011 Dataset}.
\newblock Technical Report CNS-TR-2011-001, California Institute of Technology,
  2011.

\bibitem{weitnauer2012physical}
Erik Weitnauer and Helge Ritter.
\newblock Physical bongard problems.
\newblock In {\em Ifip international conference on artificial intelligence
  applications and innovations}, pages 157--163. Springer, 2012.

\bibitem{xie2021halma}
Sirui Xie, Xiaojian Ma, Peiyu Yu, Yixin Zhu, Ying~Nian Wu, and Song-Chun Zhu.
\newblock Halma: Humanlike abstraction learning meets affordance in rapid
  problem solving.
\newblock {\em arXiv preprint arXiv:2102.11344}, 2021.

\bibitem{yi2020clevrer}
Kexin Yi, Chuang Gan, Yunzhu Li, Pushmeet Kohli, Jiajun Wu, Antonio Torralba,
  and Joshua~B. Tenenbaum.
\newblock {CLEVRER:} collision events for video representation and reasoning.
\newblock In {\em ICLR}, 2020.

\bibitem{zhuang2018hcvrd}
Bohan Zhuang, Qi Wu, Chunhua Shen, Ian~D. Reid, and Anton van~den Hengel.
\newblock {HCVRD:} {A} benchmark for large-scale human-centered visual
  relationship detection.
\newblock In Sheila~A. McIlraith and Kilian~Q. Weinberger, editors, {\em AAAI},
  2018.

\bibitem{zou2021end}
Cheng Zou, Bohan Wang, Yue Hu, Junqi Liu, Qian Wu, Yu Zhao, Boxun Li, Chenguang
  Zhang, Chi Zhang, Yichen Wei, et~al.
\newblock End-to-end human object interaction detection with hoi transformer.
\newblock In {\em CVPR}, 2021.

\bibitem{zou2021_hoitrans}
Cheng Zou, Bohan Wang, Yue Hu, Junqi Liu, Qian Wu, Yu Zhao, Boxun Li, Chenguang
  Zhang, Chi Zhang, Yichen Wei, and Jian Sun.
\newblock End-to-end human object interaction detection with hoi transformer.
\newblock In {\em CVPR}, 2021.

\end{thebibliography}
}

\end{document}